\documentclass[lettersize,journal]{IEEEtran}
\usepackage{amsmath,amsfonts}
\usepackage{algorithmic}
\usepackage{algorithm}
\usepackage{array}
\usepackage[caption=false,font=normalsize,labelfont=sf,textfont=sf]{subfig}
\usepackage{textcomp}
\usepackage{stfloats}
\usepackage{url}
\usepackage{verbatim}
\usepackage{graphicx}
\usepackage{cite}
\usepackage{tabularray}
\usepackage{multirow}
\usepackage{array}
\usepackage{ragged2e}
\usepackage{booktabs}
\usepackage{vcell}
\usepackage{svg}
\usepackage{hyperref}
\begin{document}

\title{High-Quality Exposure Correction with Diffusion-Based Image Generation Priors}

\author{Ziwen Li, Meng Cao, Jinpu Zhang, Chunyang Li, Long Bao, Heng Sun, Yuehuan Wang
    \thanks{Manuscript received XXXX \textit{(Corresponding author: Yuehuan Wang)}}
    \thanks{
        Equal Contribution: Ziwen Li, Meng Cao.

        Ziwen Li, Yuehuan Wang are with the National Key Laboratory of Science and Technology on Multispectral Information Processing, School of Artificial Intelligence and Automation, Huazhong University of Science and Technology, Wuhan 430074, China. (e-mail: D201980722@hust.edu.cn; yuehwang@hust.edu.cn)

        Meng Cao is with Mohamed bin Zayed University of Artificial Intelligence, Abu Dhabi, UAE.(email: mengcaopku@gmail.com)

        Jinpu Zhang is with the National University of Defense Technology, Changsha, China. (email: zhangjinpu@nudt.edu.cn)

        Chunyang Li, Long Bao and Heng Sun are with Xiaomi Communications Company Ltd., Beijing, China. (email: lichunyang6@xiaomi.com; baolong@xiaomi.com; hengsunheng@gmail.com)
    }
}

\markboth{Journal of \LaTeX\ Class Files,~Vol.~14, No.~8, August~2021}%
{Shell \MakeLowercase{\textit{et al.}}: A Sample Article Using IEEEtran.cls for IEEE Journals}


\maketitle

\begin{abstract}
    Although most existing exposure correction methods achieve high fidelity, they often place excessive focus on overall pixel-wise accuracy, making it challenging to effectively model extreme exposure regions, which results in suboptimal perceptual quality.
    Recently, diffusion models have received significant attention due to their remarkable performance in the realm of image generation. However, their successful application to exposure correction remains a challenging and open question. The key challenge lies in generating accurate image structures and maintaining high image fidelity during stochastic diffusion processes. 
    In this paper, we propose DPEC (Diffusion Prior-based Exposure Correction), a novel framework for image exposure correction that utilizes diffusion-based image generation priors encapsulated in pre-trained large-scale diffusion models. Specifically, we first propose an efficient fine-tuning strategy to derive an exposure corrector from pre-trained models, enabling the generation of enhanced images in a single-step denoising process. 
    Moreover, we seamlessly combine the strengths of diffusion models and regression models, and design a joint cross-attention module to integrate multi-scale diffusion prior features, thereby effectively preserving high-frequency details and minimizing random artifacts. 
    The diffusion model focuses on dealing with low-frequency content rather than all the intricate texture details.
    The experimental results demonstrate that the proposed DPEC method consistently outperforms existing state-of-the-art methods on multiple exposure correction datasets, whether in terms of fidelity, perceptual quality, or visual effects.
\end{abstract}

\section{Introduction}
Due to suboptimal exposure settings or undesirable lighting conditions, improperly exposed images may suffer from issues such as overly bright areas, excessively dark areas, or a combination of both. These exposure-related issues not only diminish the aesthetic quality of the image, causing unpleasant visual effects, but also degrade the discriminative information inherent within the image, making interpretation and analysis challenging. Exposure correction aims to recover a well-exposed image from an improperly exposed input. It serves as a critical preprocessing step for various computer vision downstream tasks, such as autonomous driving\cite{hu2023_uniad} and video understanding\cite{cao2023iterative}.

With rapid advancements of deep learning techniques, learning-based exposure correction methods have seen a leap in performance. Among these approaches, regression-based methods treat the exposure correction task as an image-to-image neural translation task, typically optimized with regression loss on paired datasets~\cite{msec,lcdpnet,sice}. Nowadays, regression-based methods have achieved significant success, especially when evaluated using distortion-based metrics such as PSNR. However, even these state-of-the-art regression methods face significant challenges in modeling well extreme exposure regions that lack sufficient structural information when reconstructing clear images. One main reason is that the regression loss equally treats each pixel in the image, focusing more on the overall pixel differences. However, extreme exposure areas usually occupy a small proportion of the entire image, which results in their smaller contribution to the total loss, leading to suboptimal perceptual quality. Consequently, how to improve the perceptual quality while maintaining the fidelity remains an urgent challenge.


Recently, diffusion models (DMs) and their successors, such as Stable Diffusion (SD), have achieved impressive performance in image generation tasks, capable of generating high-quality and diverse images. Diffusion models introduce a step-by-step denoising process, starting from pure white Gaussian noise, and gradually refining the noisy image to ultimately synthesize visually compelling results. 
During the training process, the diffusion models learn from a large amount of image data, mastering the structure, texture, and semantic information of images, enabling them to effectively generate images from scratch based on given conditional information. As a result, diffusion models possess the powerful potential to recover image content from extreme exposure areas with missing information.

Furthermore, pre-trained image diffusion models are trained on Internet-scale image collections, which provide a comprehensive encyclopedic visual representation of the real world. This comprehensive repository of visual knowledge makes it feasible to derive robust domain-specific models tailored for tasks such as exposure correction. Building on this insight, several recent approaches~\cite{StableSR,PASD} have emerged that utilize the potential of pre-trained models for image restoration by fine-tuning them to specific domains. These diffusion-based methods have shown remarkable ability to generate realistic image details.

\begin{figure*}[t]
    \centering
    \includegraphics[width = 1.0\linewidth]{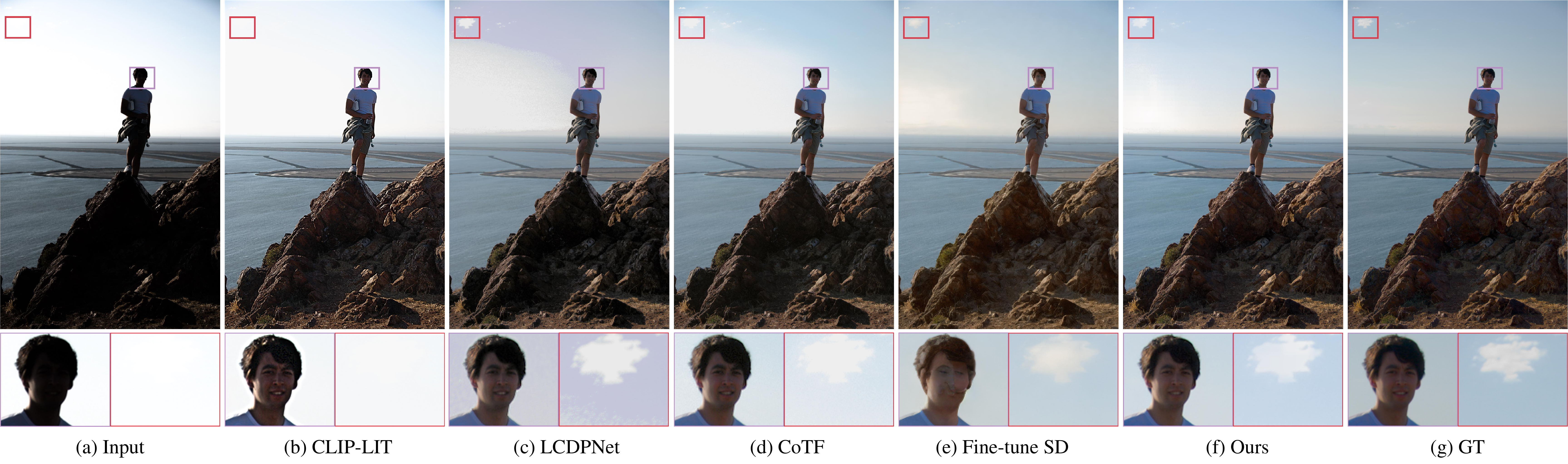}
    \caption{Comparison of visual effects. The regression-based methods, including (b), (c), and (d), may struggle to recover extreme exposure regions. Figure (e) shows the fine-tuned Stable Diffusion model, which exhibits artifacts and blurring effects in the results. Compared to these methods, our approach performs better in terms of image quality, effectively reducing artifacts and preserving details more effectively.
    }
    \label{fig:insight}
    \vspace{-4mm}
\end{figure*}

Despite significant improvements in performance, these methods are still constrained by inherent limitations for exposure correction. \textbf{1)} due to the inherent randomness of the generation process, diffusion models are prone to produce artifacts or misaligned details absent in the original input image. For instance, as illustrated in Fig.~\ref{fig:insight}, even after fine-tuning SD using degraded images as input conditions, the generated outputs still exhibit discordant elements, such as pronounced artifacts in facial and hand regions. \textbf{2)} The iterative generation process of diffusion models is computationally intensive, typically requiring a substantial number of sampling steps. Although advanced sampling strategies have been proposed to mitigate this issue, the overall computational cost remains significantly high. \textbf{3)} In order to process high-resolution images, SD employs a VAE to compress the image into a latent space to reduce the computational and memory demands. However, the high compression rate (e.g., 1/8) of the VAE leads to the loss of considerable high-frequency details, resulting in severe blurring in the reconstructed images. For exposure correction tasks, it is essential that the model produce high-quality, high-resolution enhanced images within a reasonable timeframe while preserving fine details and avoiding discordant artifacts. Nonetheless, the aforementioned challenges lead to suboptimal performance when fine-tuning diffusion models with existing strategies.

To overcome the above barriers, in this work, we propose a novel Diffusion Prior-based Exposure Correction (DPEC) method, which leverages the powerful diffusion-based image generation priors encapsulated in a pre-trained large-scale image diffusion model. We first present a simple and efficient fine-tuning strategy to transform a pre-trained diffusion model into an exposure correction model. Specifically, the input to the denoising UNet is modified from pure noise to a combination of degraded images and noise maps. Since the input already contains most of the pixel structure of the expected image, the denoising pathway can be greatly streamlined. By focusing on a specific level of denoising, our method enables high-quality image generation in just a single step, thus reducing inference time. 
Furthermore, given the advantages of regression-based methods in terms of fidelity, we integrate diffusion models with regression-based models to preserve image details and avoid artifacts. Unlike previous approaches that rely on traditional backbones for vanilla feature extraction, we seamlessly integrate generators and regressors as image backbones. 
Specifically, low-resolution components are fed into the fine-tuned denoising UNet, and multi-scale diffusion priors are hierarchically integrated into the regression model through a proposed joint cross-attention module. 
In this way, by focusing on recovering the low-frequency content during the diffusion process, we fully exploit the generative benefits of diffusion model and minimize random artifacts. Simultaneously, the regression model effectively preserves the high-frequency details. Equipped with the above design, our method consistently reconstructs high-quality enhanced results. Extensive experiments on several exposure correction datasets confirm that our proposed DPEC outperforms state-of-the-art methods in terms of fidelity, perceptual quality, and visual effect.

The main contributions are summarized as follows:
\begin{itemize}
    \item We delve into the utilization of pre-trained diffusion models to achieve high-quality exposure correction and propose a novel Diffusion Prior-based Exposure Correction (DPEC) method.
    \item We propose a simple and efficient fine-tuning strategy to derive an exposure correction model that requires only a single inference step.
    \item We seamlessly integrate the advantages of diffusion models and regression models, avoiding artifacts while preserving details. Additionally, we design a joint cross-attention module to hierarchically integrate diffusion prior features.
    \item We validate the effectiveness of the proposed method on multiple datasets. Extensive experiments demonstrate that the proposed method achieves competitive performance in terms of fidelity, perceptual quality, and visual effects.
\end{itemize}

\section{Related Work}
\subsection{Exposure Correction}

\subsubsection{Traditional Methods}
Traditional exposure correction methods typically rely on hand-crafted priors to adjust image contrast, employing techniques such as histogram equalization~\cite{he,clahe,generalizations_HE}, curve adjustment~\cite{bennett2005video,yuan2012automatic} and Retinex-based decomposition~\cite{SSR}. Histogram equalization redistributes pixel intensities for more uniform brightness, while curve adjustment utilizes predefined nonlinear mappings (e.g., gamma correction) to adjust brightness. Retinex-based methods decompose images into illumination and reflectance components and enhance them separately. Wang et al.~\cite{NPE} applied a bright-pass filter with double logarithmic transformation for non-uniform illumination. Li et al.~\cite{robust_retinex} incorporated noise maps to improve robustness, and Fu et al.~\cite{WVM} developed a weighted variational model to preserve details and suppress noise. Zhang et al.~\cite{zhang2018high} introduced perceptual similarity constraints for illuminance map estimation, and Zhang et al.~\cite{zhang2019dual} proposed a dual illuminance estimation method addressing both underexposure and overexposure. Despite their effectiveness in specific scenarios, these methods often depend heavily on hand-crafted priors, leading to suboptimal performance in complex scenes.

\subsubsection{Deep Learning-based Methods}
Recently, deep learning-based methods have achieved impressive performance. 
Some methods focus on underexposed image enhancement and can be divided into two main categories.
One class of methods~\cite{ZeroShotRetinex,self_reinforced,lcdpnet,Untrained,pairlie,li2024rethinking,liu2024pixel} combines deep neural networks with physical models in a data-driven manner. For instance, RetinexNet~\cite{retinexnet,retinexnet2} and KinD~\cite{KinD,kind_plus} employ multilevel sub-networks for Retinex decomposition, light enhancement and reflection denoising. RUAS~\cite{RURS} utilizes a collaborative search strategy to discover the optimal compact Retinex-inspired network. URetinexNet~\cite{uretinex} unfolds the optimization process of Retinex decomposition through deep neural networks. HDRNet~\cite{HDRNet} predicts local affine transformation coefficients with learnable bilateral grid up-sampling for real-time enhancement. DeepUPE~\cite{DeepUPE} refines light map estimation to correlate inputs with enhanced results. SCI~\cite{sci} proposed a cascading process to learn an illumination map, and Zero-DCE~\cite{ZeroDCE,zerodce_plus} iteratively estimates light enhancement curves to improve the dynamic range.
Another class of methods directly learn image-to-image mappings through end-to-end learning, adopting various techniques such as Laplacian pyramid~\cite{luminancePyramid,Laplacian}, wavelet transform~\cite{wavelet,half_wavelet,attention-frequency}, generative adversarial network~\cite{EnlightenGAN,Adversarial1,DPE,corss-disentanglement,DRBN2}, normalized flow~\cite{flow}, Transformer~\cite{snr_transformer,ultra_transformer,xu2024bilateral} and semantic guidance~\cite{li2024sam,wu2023_semantic}. Yang et al.~\cite{DRBN} proposed a semi-supervised learning framework to improve fidelity and perceptual quality. Wu et al.~\cite{wu2023_semantic} used semantic segmentation maps to guide enhancement, and SNRNet~\cite{snr_transformer} introduces an SNR-guided self-attention for spatially varying dynamic enhancement.

Recently, some methods have started to focus on both overexposure and underexposure issues, aligning more closely with practical applications.
Afifi et al.~\cite{msec} designed a coarse-to-fine multi-scale network and proposed a large-scale exposure correction dataset. Huang et al.~\cite{enc,huang2022exposure} proposed to learn exposure-invariant feature spaces to narrow the gap between different exposure levels. FECNet~\cite{fecnet} decomposes and reconstructs the brightness and structure components within the Fourier domain and incorporates spatial interactions. Wang et al.~\cite{wang2023decoupling} proposed frequency domain decomposition in convolution to enhance contrast and detail. CuDi~\cite{cudi} designed the curve distillation technique to approximate the iterative processes of the traditional curve framework to improve the speed of inference. Wang et al.~\cite{lcdpnet} introduced local color distributions to tackle non-uniform illumination. Huang et al.~\cite{huang2023learning} addressed optimization conflicts by learning the sample relation within the mini-batch. CLIP-LIT~\cite{cliplit} proposed to utilize the CLIP to learn initial prompt pairs and train unsupervised enhancement networks. CoTF~\cite{CoTF} proposes a collaborative transformation framework and adaptive sampling strategies for real-time exposure correction. Li et al.~\cite{OSMamba} proposed an omnidirectional spectral Mamba architecture with a dual-domain prior generator for exposure correction. Huang et al.~\cite{CLIPRestoreX} leveraged CLIP to jointly restore image structure and perceptual quality in exposure correction.

However, most methods utilize a regression-based learning paradigm, which tends to place under emphasis on fidelity metrics while neglecting the importance of perceptual quality. In contrast, our method seamlessly integrates diffusion models and regression models, leveraging the generation priors encapsulated in the pre-trained diffusion models to achieve superior performance.

\subsection{Diffusion Models}

Denoising Diffusion Probabilistic Models (DDPMs)~\cite{DDPM} have emerged as a highly effective generative framework, demonstrating state-of-the-art performance in image synthesis. DDPM consists of a forward process that gradually adds noise and a backward process that removes it.
DDIMs~\cite{DDIM} extend DDPMs by introducing non-Markovian paths that significantly reduce the number of sampling steps while maintaining high-quality synthesis. Conditional diffusion models~\cite{DiffusionBeatGAN,CFG} extend this framework by incorporating additional conditional information to guide the generation process, facilitating controlled generation and expanding the scope of potential applications.
Rombach et al. developed a Stable Diffusion (SD)~\cite{LDM} that operates within the latent space of VAE compression. 
Trained on the large-scale LAION-5B dataset, this model encodes an extensive collection of Internet images into its weights, endowing it with rich priors for natural image generation. As a result, the SD is capable of producing a diverse range of high-quality images.
Recently, many methods have emerged that leverage pre-trained generative priors for image restoration tasks, e.g., StableSR~\cite{StableSR}, DiffBIR~\cite{lin2024diffbir} and PASD~\cite{PASD}. These methods typically fine-tune pre-trained Stable Diffusion models by means of ControlNet~\cite{ControlNet}, which has demonstrated an excellent ability to generate realistic image details. ExposureDiffusion~\cite{ExposureDiffusion} introduced a diffusion-based framework that models the exposure process as a reverse diffusion for image enhancement. Diff-Retinex~\cite{yi2023diff} combined Retinex decomposition with generative diffusion models, utilizing the diffusion process to refine illumination and reflectance components. 

Beyond 2D image tasks, there is a growing interest in leveraging diffusion priors for more complex applications, including 3D/4D reconstruction, scene generation, and physical modeling. For instance, DimensionX~\cite{sun2024dimensionx} uses video diffusion models to provide spatio-temporal priors for 3D/4D reconstruction, VideoScene~\cite{wang2025videoscene} distills video diffusion models for 3D scene generation, ReconX~\cite{liu2024reconx} employs video diffusion priors for sparse-view reconstruction, and Physic3D~\cite{liu2024physics3d} distills physical priors from video diffusion models. These works demonstrate the potential of diffusion priors to provide structured guidance beyond standard image synthesis tasks.

However, exposure correction imposes high demands on preserving image details and avoiding artifacts. Given the randomness of diffusion models and their high computational cost, how to effectively leverage the prior knowledge embedded in pre-trained diffusion models for image generation for exposure correction remains a significant challenge. In this study, we delve into the potential of achieving high-quality exposure correction based on pre-trained diffusion models, providing new perspectives for this field.

\begin{figure*}[t]
    \centering
    \includegraphics[width = 1.0\textwidth]{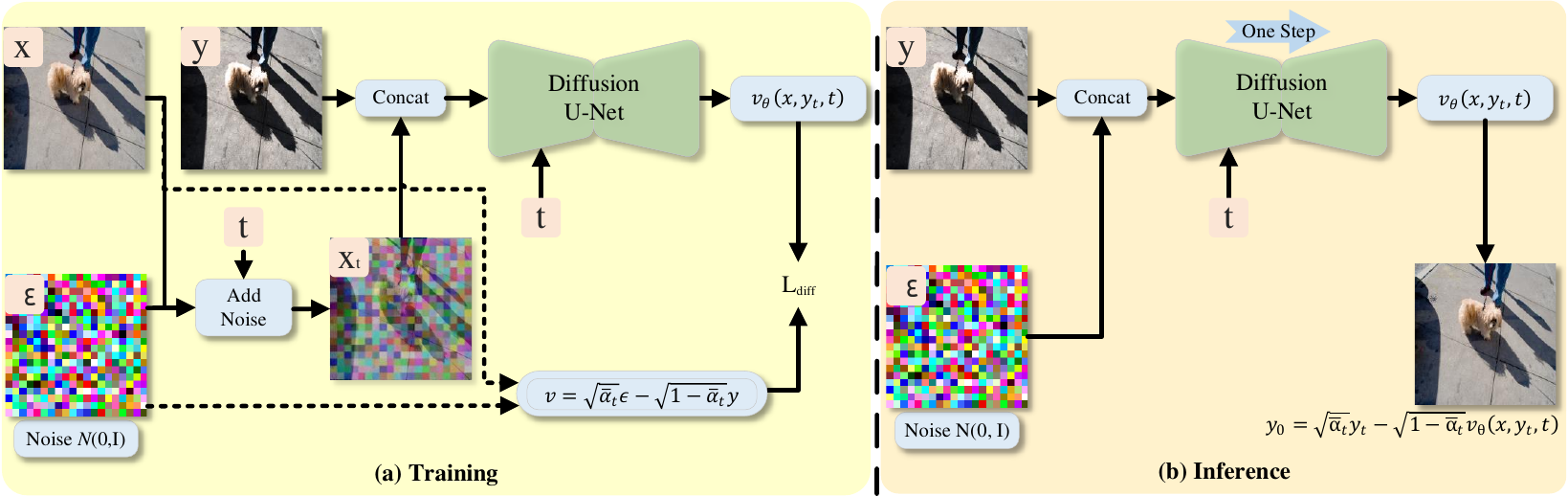}
    \caption{The diagram illustrating the training and inference process of the fine-tuning strategy. Starting from the pre-trained stable diffusion model, we achieve conditional control by concatenating images with noise and inputting them into the U-Net. During the training process, we fine-tune the parameters of the U-Net using fixed time steps and optimize through the v-predicted objective function. In the inference process, we use single-step sampling to generate images.}
    \label{fig:finetune}
    \vspace{-4mm}
\end{figure*}

\section{Method}
In this section, we first review the theoretical background of diffusion models. Then, we describe the detailed workflow of DPEC, including the fine-tuning strategy and the architecture design. Finally, we introduce the two-stage training strategy.

\subsection{Preliminaries}  
Denoising diffusion probabilistic models (DDPMs) are a class of latent variable models based on Markov chains. The core idea is to gradually diffuse the complex data distribution to a simple prior distribution (e.g., Gaussian distribution) and then learn the inverse process by training a parameterized network.
In order to control the direction of generation, the conditional diffusion generative model trains a neural network to model the conditional distribution $P(y|x)$, where the condition $x$ represents the conditional information, and $y$ corresponds to the target distribution.

Diffusion models consist of two processes: a forward process that progressively adds noise to the data, and an inverse process that iteratively predicts denoised samples. The forward process starts with a given data distribution $y_0 \sim q(y)$. By progressively adding Gaussian noise, it generates a noisy sample \( y_t \) at each timestep \( t \). This process can be mathematically expressed as:
\begin{align}
    \mathbf{y}_t = \sqrt{\bar{\alpha}_t} \mathbf{y}_0 + \sqrt{1 - \bar{\alpha}_t} \boldsymbol{\epsilon}
\end{align}
where $\{\alpha_t\}_{t=1}^T$ is the noise schedule, and $\epsilon \sim \mathcal{N}(0, \mathbf{I})$ represents Gaussian noise sampled from a standard normal distribution.

Conversely, the reverse process aims to reconstruct the original data from its ultimate noisy state. Beginning with $y_T \sim N(0, \mathbf{I})$, the process iteratively refines the data, progressively transforming the noisy variable $y_{t}$ into a cleaner version $y_{t-1}$ at each timestep $t$.
This is achieved by the denoising model $\epsilon_\theta$ with learned parameters $\theta$. The objective function of the training process $L_{\text{MSE}}$ aims to make the generated data approximate the true data distribution, which is formulated as
\begin{align}
    L_{\text{MSE}} = \mathbb{E}_{t,x,\epsilon \sim \mathcal{N} (0,\mathbf{I})} \left\| \epsilon_\theta(y_t, t, x) - \epsilon \right\|^2_2,
\end{align}
where $t$ is uniformly sampled from $\{1, . . . , T \}$.

Nevertheless, the implementations of numerous leading methods remain unavailable to the public. Stable Diffusion is the first publicly available large-scale model, significantly contributing to the widespread application of image synthesis. It provides rich prior knowledge for image generation, which can be further fine-tuned to meet specific domain requirements.

\subsection{Fine-Tuning Strategy}
Training diffusion models from scratch demands substantial data and computational resources, so we leverage pre-trained Stable Diffusion models as a foundation. In this work, we formulate exposure correction as a conditional denoising diffusion generation problem and use the degraded input image as conditional information. Building on this, we propose an efficient fine-tuning strategy that adapts the model to exposure correction tasks with minimal modifications to the model components. Fig.~\ref{fig:finetune} illustrates the pipeline of our fine-tuning strategy.

\subsubsection{Architecture Modification}
The core objective of VAE training is to achieve compression and reconstruction of high-resolution images. However, VAE reconstruction may lose high-frequency details and introduce artifacts. Considering the additional computational overhead brought by VAE, we chose to remove it, directly using downsampled images as input and relying on a regression model to ensure the recovery of high-frequency details.
To further simplify the model, we disable the text conditioning by setting the text encoding to an empty string. Although additional adapters are often used to incorporate conditional guidance, they increase model complexity. 
To control the generation direction with minimal architectural modifications, we propose concatenating the downsampled image $x$ and the noisy image $y_t$ along the channel dimension as inputs to the network, represented as $\hat{y} = [y_t, x]$.
Consequently, the input convolution layer of the UNet is modified to accommodate 6 input channels, corresponding to the increased dimensionality.
To gradually integrate conditional information, we adopt zero convolution weight initialization, where the convolution layer weights corresponding to the condition \(x\) are initialized to zero.
In addition, considering the impact of resolution on the signal-to-noise ratio of noisy images, we select a fixed input resolution of \(64 \times 64\), consistent with the original Stable Diffusion model.

\begin{figure*}[t]
    \centering
    \includegraphics[width = 1.0\textwidth]{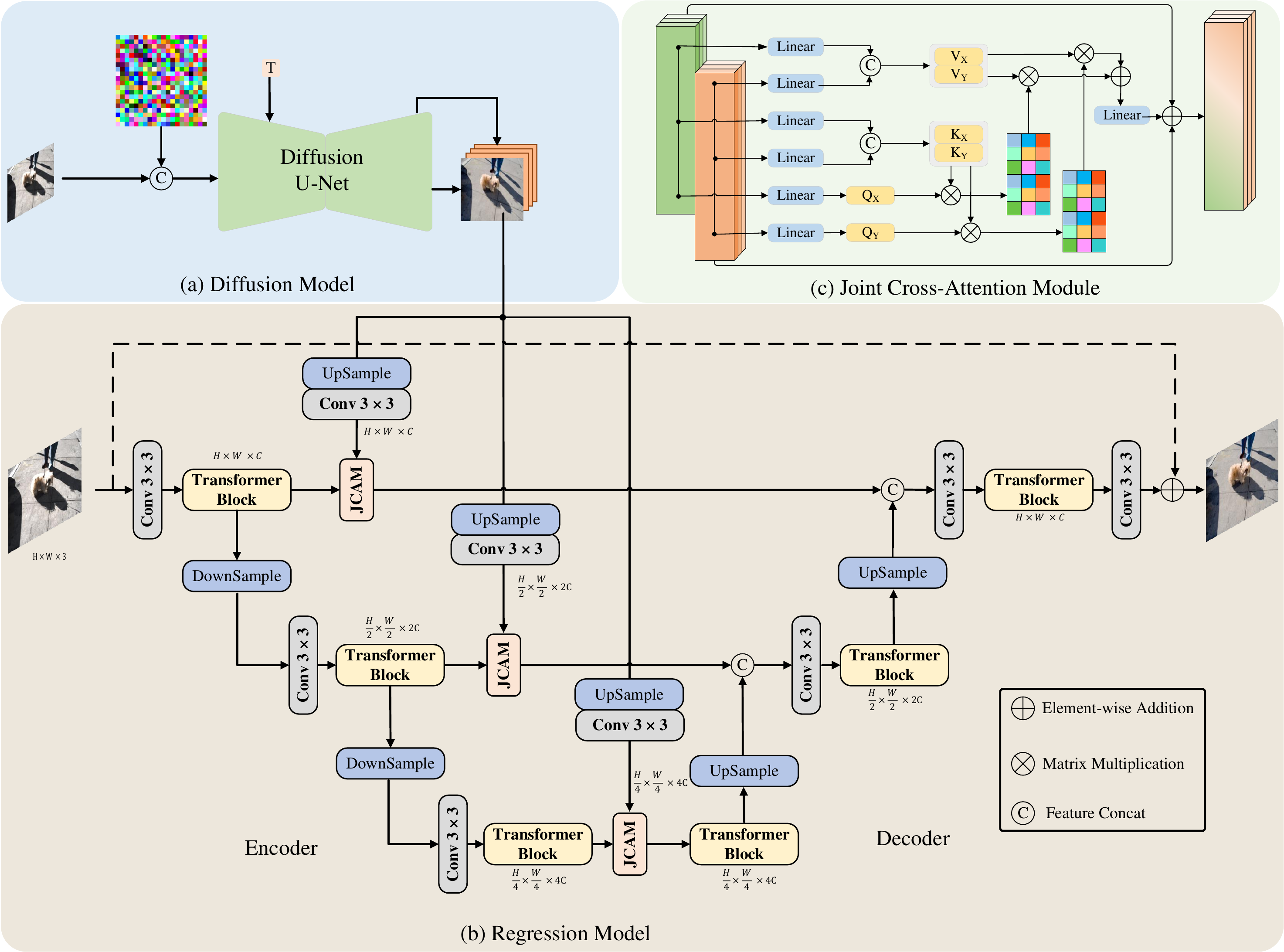}
    \caption{Our DPEC overall framework:
    (a) Diffusion model: Perform single-step denoising in the low-resolution space.
    (b) Regression model: Use a hierarchical encoder-decoder structure and effectively integrate multi-scale diffusion prior features.
    (c) Joint Cross-Attention Module (JCAM): Compute joint cross-attention between diffusion features and regression features.}
    \label{fig:framework}
    \vspace{-4mm}
\end{figure*}

\subsubsection{Single-Step Generation}  
Most existing text-guided diffusion models generate almost every pixel from scratch and typically adopt a step-by-step iterative approach to gradually approximate the target image. However, the exposure correction task is quite different. The input image $\hat{y}$ contains the poorly exposed image that already provides the basic pixel structure of the expected image. Therefore, the target image can be approximated more efficiently during the generation process. 
In the DDIM sampling process, we found that a single sampling step is sufficient to generate high-quality and clear images, without relying on the multi-step diffusion process typically used in text-to-image generation. Furthermore, increasing the number of time steps provides almost no additional improvement in the quality of the generated images.
Based on this observation, we further explored how to better utilize input information, reduce redundant calculations and unnecessary randomness, and achieve efficient generation through single-step sampling.


We argue that denoising UNet is a network specifically designed to remove a wide range of Gaussian noises, and can adaptively adjust its denoising strategy based on the intensity of the input noise. However, having a single network learn to remove multiple intensities of noise (e.g., the common 1000) would have a significant learning burden. To reduce the complexity of learning, we choose to focus on denoising tasks with specific intensity levels. Doing so helps the model to understand and process this specific level of noise more accurately, thus allowing the model to generate high-quality images in just a single step. Since different timesteps correspond to different noise intensities, we fixedly choose a timestep $t$ during the training process. To maintain consistency between training and testing, we use the maximum timestep $T$ during training. With this way, the model is directed to focus on a single intensity of noise removal, which improves denoising effectiveness and inference efficiency.

\subsubsection{Parameterization}
We parameterize the UNet using the \textit{v-prediction} strategy, which works well in image generation tasks. Specifically, the UNet model is fine-tuned to use the diffusion loss $L_{diff}$, which is expressed as
\begin{align}
    L_{diff} = \mathbb{E}_{x,y,t,\epsilon} \left[ \left\| \left( \sqrt{\bar{\alpha}_t} \epsilon - \sqrt{1 - \bar{\alpha}_t} y \right) - v_\theta(x, y_t, t) \right\|_2^2 \right],
\end{align}
where $t$ is set to $T$, $v_\theta(x, y_t, t)$ is the UNet prediction and $y$ is the ground truth image.
During the sampling stage, after getting $v_\theta(x, y_t, t)$, we can directly get the generated image by calculating:
\begin{align}
    y_0 = \sqrt{\bar{\alpha}_t} y_t - \sqrt{1 - \bar{\alpha}_t} v_{\theta} (x, y_t, t),
\end{align}
where $y_0$ is the enhanced result , $t$ is $T$ and $y_T \sim N(0, \mathbf{I})$.

\subsection{Multi-Scale Integration Architecture}
Given the advantages of regression models in terms of fidelity and the effectiveness of multi-scale design, we introduce a multi-scale integration architecture, as shown in Fig.~\ref{fig:framework}. The goal is to effectively fuse the diffusion prior feature with regression models to obtain better fidelity and perceptual quality.

For a given image \( I \in \mathbb{R}^{h \times w \times 3} \), we first downsample it to obtain a fixed-resolution version \( I_{low} \), which is then processed by the diffusion model. During the downsampling process, high-frequency details are lost, leaving \( I_{low} \) to primarily retain low-frequency information. Consequently, the diffusion model generates image content that is dominated by low-frequency components. 
Specifically, we first use the fine-tuned denoising UNet to process \( I_{low} \), producing a result \( \hat{I_{low}} \) with high perceptual quality. Since the low-resolution image inherently contains fewer details, this effectively prevents the introduction of unnecessary artifacts during generation. Then, we concatenate the features from the last layer of the UNet with the generated image \( \hat{I_{low}} \) to construct the diffusion prior features \( X \).

Next, we feed the original resolution image \( I \) into an encoder-decoder structure. Both the encoder and decoder are composed of multiple Transformer blocks operating along the channel dimension, efficiently capturing global contextual feature \( Y \). This process is conducted at the original resolution, enabling better preservation of high-frequency details. Additionally, we propose a joint cross-attention module to further fuse diffusion prior features across multiple scales. By integrating hierarchical diffusion priors, our approach seamlessly combines the advantages of both diffusion and regression models. The diffusion model generates missing content, while the regression model refines high-frequency structures. This complementary synergy leads to remarkable improvements in both fidelity and perceptual quality.

\subsection{Joint Cross-Attention Module} 
After extracting the diffusion features $X$ and regression features $Y$, it is crucial to fuse these two complementary features. $X$ has good perceptual quality, while $Y$ contains fine image details. Previous methods often adopt a cross-attention mechanism, which is an asymmetric fusion strategy where conditional features are only used to compute attention weights without directly participating in the weighted summation. This design may limit the ability to fully integrate information from both features, leading to suboptimal results.

To this end, we design a Joint Cross-Attention Module (JCAM), as depicted in Fig.~\ref{fig:framework}(c). The JCAM enables the interaction of two features during the computation of attention weights and the generation of fused features. This ensures symmetric information transmission and thorough interaction, achieving a more comprehensive fusion of both features.
Specifically, for given features \( X \) and \( Y \), we first perform layer normalization on them separately. Then, we transform them into query (Q), key (K), and value (V) matrices. The transformation formulas are as follows:
\begin{align}
    Q_X = W_Q^X X, \quad K_X = W_K^X X, \quad V_X = W_V^X X, \\
    Q_Y = W_Q^Y Y, \quad K_Y = W_K^Y Y, \quad V_Y = W_V^Y Y,
\end{align}
where \( W_Q^X, W_K^X, W_V^X, W_Q^Y, W_K^Y, W_V^Y \) are linear transformation matrices.
Next, we concatenate the key and value matrices of features \( X \) and \( Y \) to obtain:
\begin{align}
    K_{\text{cat}} &= [K_X, K_Y], \\
    V_{\text{cat}} &= [V_X, V_Y].
\end{align}
We then compute the attention weight matrices \( A_X \) and \( A_Y \) by conditioning on \( Q_X \) and \( Q_Y \) with the concatenated key matrix \( K_{\text{cat}} \), which is formulated as
\begin{align}
    A_X = \text{softmax}\left(\frac{Q_X K_{\text{cat}}^T}{\sqrt{C}}\right), \\
    A_Y = \text{softmax}\left(\frac{Q_Y K_{\text{cat}}^T}{\sqrt{C}}\right),
\end{align}
where \( C \) is the dimension of the key matrix, acting as a scaling factor.
To reduce computational burden, we perform cross-attention calculations along the channel dimension rather than the spatial dimension. Finally, we use the attention weight matrices to compute the weighted sum of features, obtaining \( F_X \) and \( F_Y \):
\begin{align}
    F_X = A_X V_{\text{cat}}, \\
    F_Y = A_Y V_{\text{cat}}.
\end{align}
Subsequently, we combine these two features and apply a linear projection to obtain the final output. Finally, we enhance the feature representation through a feedforward network. JCAM effectively integrates diffusion features and regression features, thereby better maintaining fidelity and perceptual quality.

\subsection{Training}
We adopt a two-stage training strategy to train DPEC, which involves fine-tuning the Stable Diffusion and training the regression network. In the first stage, we fine-tune the denoising UNet under the guidance of \(L_{\text{diff}}\) using proposed fine-tuning strategy.

In the second stage, we fix and freeze the parameters of the UNet, then train the network to learn regression-based image reconstruction capabilities. Consistent with previous works~\cite{enc}, we use regression loss $L_{reg}$ to reconstruct well-exposed images. The regression loss consists of the L1 loss $L_{1}$, the perceptual loss $L_{per}$, and the SSIM loss $L_{ssim}$. The loss function is defined as
\begin{align}
    L_{reg} =  L_{1} + \alpha L_{per} + \beta L_{ssim}
\end{align}
where $\alpha$ and $\beta$ are the weight coefficients of the respective losses.

\subsection{Discussion}
\textbf{Single-step sampling.} Our single-step sampling strategy is further optimized based on the DDIM sampling method. As a general sampling strategy, DDIM typically requires multiple iterative denoising steps to achieve high-quality generation. However, in our experiments, we observed that for tasks such as exposure correction, where low-quality images serve as conditions, the input itself already provides most of the structural information of the target image. In this case, the model only needs to refine illumination and details during sampling, rather than reconstructing the image step by step from noise as in generic generation tasks. Therefore, single-step sampling is sufficient to produce high-quality results, while multi-step sampling does not bring significant improvements. Moreover, to further reduce the learning difficulty, we focus on a fixed noise level during training instead of training across multiple noise levels. This design effectively alleviates the model's learning burden and allows it to concentrate more on the optimization objectives of the task itself. In summary, our approach is a task-specific optimization of DDIM tailored for exposure correction. By combining single-step sampling with fixed noise, we significantly improve inference efficiency while maintaining generation quality, and we simplify the training process, making it more efficient and better aligned with the practical requirements of exposure correction.

\textbf{Integration of diffusion and regression models.} Unlike StableSR\cite{StableSR}, our method introduces a regression model to ensure higher fidelity. Specifically, simply fine-tuning Stable Diffusion is insufficient to guarantee realistic results. On the one hand, diffusion models may generate artifacts or misaligned details; on the other hand, the high compression rate of the VAE makes it difficult to preserve high-frequency details, leading to blurry results. Therefore, directly fine-tuning Stable Diffusion is not ideal for exposure correction tasks. To address this, our method assigns Stable Diffusion to process only the low-frequency components of the image, while the high-frequency details are entirely preserved by the regression model.

\textbf{Difference from existing diffusion-based exposure correction methods.} Compared with existing diffusion-based exposure correction methods, our approach differs significantly in several aspects.
In terms of leveraging the diffusion prior, \cite{yi2023diff} only uses the diffusion model for Retinex reconstruction, without fully exploiting the generative prior. In contrast, our method leverages the diffusion prior to more effectively constrain the output distribution, leading to reconstructions with improved fidelity and perceptual quality.
Regarding the learning paradigm, \cite{lv2024fourier} employs unsupervised learning, performing luminance correction via Fourier domain decomposition and guiding image restoration with unsupervised loss; \cite{cho2024zero} relies on zero-shot conditional generation, producing results based on conditional prompts but lacking explicit supervisory signals. By comparison, our method adopts supervised learning, using paired data and well-defined loss functions to directly optimize the target image quality.
Concerning the integration of diffusion and regression models, \cite{lv2024fourier,cho2024zero} rely solely on diffusion models, while \cite{li2024sagiri} follows a regression-then-diffusion strategy, which limits detail preservation and results in suboptimal enhancement. Our two-stage strategy seamlessly combines diffusion and regression: it first leverages the diffusion model to generate a high-quality initial representation, and then integrates the diffusion prior into a regression model. This design effectively preserves fine high-frequency details while maintaining global consistency.
Furthermore, in terms of fine-tuning and inference efficiency, existing methods typically rely on multi-step sampling, which is computationally expensive. To address this, we design a single-step sampling strategy that significantly improves inference speed while maintaining enhancement quality.

\section{Experiments}
\subsection{Experimental Settings}
\subsubsection{Datasets}
To evaluate the performance of our proposed method, we conducted experiments on three benchmark datasets: LCDP~\cite{lcdpnet}, MSEC~\cite{msec} and SICE~\cite{sice}. Specifically, the LCDP dataset primarily contains scenes with non-uniform illumination, featuring both overexposed and underexposed regions within a single image. It comprises 1415 training images, 100 validation images, and 218 test images. The MSEC dataset contains multi-exposure scenes with various exposure levels, including 17,675 training images, 750 validation images, and 5,905 test images. For the SICE dataset, following the setup in~\cite{enc}, we use the second and the penultimate exposure subsets to represent overexposed and underexposed images, respectively. The middle exposure subset serves as the reference images. The SICE dataset consists of 1,000 image pairs for training and 60 image pairs for testing. Additionally, we also evaluate the generalization performance using the unlabeled LIME, DICM, and NPE datasets.

\begin{table*}
\centering
\caption{Quantitative Comparison on MSEC and SICE Datasets in Terms of PSNR, SSIM, LPIPS, and NIQE. The Best Results Are Highlighted in Bold.}
\label{table:msec_sice}
 \resizebox*{\linewidth}{!}{
\begin{tabular}{c!{\vrule width \lightrulewidth}cc!{\vrule width \lightrulewidth}cc!{\vrule width \lightrulewidth}cccc!{\vrule width \lightrulewidth}cc!{\vrule width \lightrulewidth}cc!{\vrule width \lightrulewidth}cccc} 
\toprule
\multirow{3}{*}{Methods} & \multicolumn{8}{c!{\vrule width \lightrulewidth}}{MSEC}                                                                                                                                 & \multicolumn{8}{c}{SICE}                                                                                                                                                                 \\ 
\cmidrule[\heavyrulewidth]{2-17}
                         & \multicolumn{2}{c!{\vrule width \lightrulewidth}}{Under} & \multicolumn{2}{c!{\vrule width \lightrulewidth}}{Over} & \multicolumn{4}{c!{\vrule width \lightrulewidth}}{Average}         & \multicolumn{2}{c!{\vrule width \lightrulewidth}}{Under} & \multicolumn{2}{c!{\vrule width \lightrulewidth}}{Over} & \multicolumn{4}{c}{Average}                                         \\
                         & PSNR           & SSIM                                    & PSNR           & SSIM                                   & PSNR           & SSIM            & LPIPS           & NIQE          & PSNR           & SSIM                                    & PSNR           & SSIM                                   & PSNR           & SSIM            & LPIPS           & NIQE           \\
\midrule
HE~\cite{he}                        & 16.52          & 0.6918                                  & 16.53          & 0.6991                                 & 16.53          & 0.6959          & 0.2920          & 3.74          & 14.69          & 0.5651                                  & 12.87          & 0.4991                                 & 13.78          & 0.5376          & 0.3738          & 3.49           \\
CLAHE~\cite{clahe}                    & 16.77          & 0.6211                                  & 14.45          & 0.5842                                 & 15.38          & 0.5990          & 0.4744          & 3.87          & 12.69          & 0.5037                                  & 10.21          & 0.4847                                 & 11.45          & 0.4942          & 0.4688          & 3.82           \\
LIME~\cite{LIME}                     & 13.98          & 0.6630                                  & 9.88           & 0.5700                                 & 11.52          & 0.6070          & 0.2758          & 3.67          & 16.48          & 0.5832                                  & 6.67           & 0.4041                                 & 11.58          & 0.4937          & 0.3712          & 3.26           \\
WVM~\cite{WVM}                       & 18.67          & 0.7280                                  & 12.75          & 0.6450                                 & 15.12          & 0.6780          & 0.2284          & 3.68          & 15.16          & 0.5915                                  & 8.03           & 0.4485                                 & 11.60          & 0.5200          & 0.3432          & 3.10           \\
RetinexNet~\cite{retinexnet}               & 12.13          & 0.6209                                  & 10.47          & 0.5953                                 & 11.14          & 0.6048          & 0.3209          & 4.06          & 12.94          & 0.5171                                  & 12.87          & 0.5252                                 & 12.90          & 0.5212          & 0.4312          & 3.52           \\
URetinexNet~\cite{uretinex}              & 13.85          & 0.7371                                  & 9.81           & 0.6733                                 & 11.42          & 0.6988          & 0.2858          & 3.73          & 17.39          & 0.6448                                  & 7.40           & 0.4543                                 & 12.40          & 0.5496          & 0.3549          & 3.50           \\
DRBN~\cite{DRBN}                     & 19.74          & 0.8290                                  & 19.37          & 0.8321                                 & 19.52          & 0.8309          & 0.2795          & 3.91          & 17.96          & 0.6767                                  & 17.33          & 0.6828                                 & 17.65          & 0.6798          & 0.3891          & 3.53           \\
SID~\cite{SID}                       & 19.37          & 0.8103                                  & 18.83          & 0.8055                                 & 19.04          & 0.8074          & 0.1862          & 3.74          & 19.51          & 0.6635                                  & 16.79          & 0.6444                                 & 18.15          & 0.6540          & 0.2417          & 3.70           \\
MSEC~\cite{msec}                     & 20.52          & 0.8129                                  & 19.79          & 0.8156                                 & 20.08          & 0.8145          & 0.1721          & 3.68          & 19.62          & 0.6512                                  & 17.59          & 0.6560                                 & 18.58          & 0.6536          & 0.2814          & \textbf{2.88}  \\
Zero-DCE~\cite{ZeroDCE}                   & 14.55          & 0.5887                                  & 10.40          & 0.5142                                 & 12.06          & 0.5441          & 0.2923          & 3.74          & 16.92          & 0.6330                                  & 7.11           & 0.4292                                 & 12.02          & 0.5311          & 0.3532          & 3.24           \\
Zero-DCE++~\cite{zerodce_plus}               & 13.82          & 0.5887                                  & 9.74           & 0.5142                                 & 11.37          & 0.5583          & 0.3121          & 3.77          & 11.93          & 0.4755                                  & 6.88           & 0.4088                                 & 9.41           & 0.4422          & 0.3623          & 3.27           \\
RUAS~\cite{RURS}                     & 13.43          & 0.6807                                  & 6.39           & 0.4655                                 & 9.20           & 0.5515          & 0.4819          & 5.91          & 16.63          & 0.5589                                  & 4.54           & 0.3196                                 & 10.59          & 0.4393          & 0.5122          & 6.79           \\
SCI~\cite{sci}                      & 9.97           & 0.6681                                  & 5.83           & 0.5190                                 & 7.49           & 0.5786          & 0.3116          & 3.73          & 17.86          & 0.6401                                  & 4.45           & 0.3629                                 & 12.49          & 0.5051          & 0.4239          & 3.54           \\
PairLIE~\cite{pairlie}                  & 11.78          & 0.6596                                  & 8.37           & 0.5887                                 & 9.73           & 0.6171          & 0.3605          & 4.27          & 16.67          & 0.5995                                  & 6.26           & 0.3846                                 & 11.47          & 0.4921          & 0.4138          & 3.96           \\
ENC-SID~\cite{enc}                  & 22.59          & 0.8423                                  & 22.36          & 0.8519                                 & 22.45          & 0.8481          & 0.1827          & 4.53          & 21.30          & 0.6645                                  & 19.63          & 0.6941                                 & 20.47          & 0.6793          & 0.2797          & 3.47           \\
ENC-DRBN~\cite{enc}                 & 22.72          & 0.8544                                  & 22.11          & 0.8521                                 & 22.35          & 0.8530          & 0.1724          & 4.61          & 21.77          & 0.7052                                  & 19.57          & 0.7267                                 & 20.67          & 0.7150          & 0.2318          & 3.33           \\
CLIP-LIT~\cite{cliplit}                 & 17.79          & 0.7611                                  & 12.02          & 0.6894                                 & 14.32          & 0.7181          & 0.2506          & 3.58          & 15.13          & 0.5847                                  & 7.52           & 0.4383                                 & 11.33          & 0.5115          & 0.3560          & 3.49           \\
FECNet~\cite{fecnet}                   & 22.96          & 0.8598                                  & 23.22          & 0.8748                                 & 23.12          & 0.8688          & 0.1419          & 3.78          & 22.01          & 0.6737                                  & 19.91          & 0.6961                                 & 20.96          & 0.6849          & 0.2656          & 3.58           \\
LCDPNet~\cite{lcdpnet}                  & 22.35          & \textbf{0.8650}                         & 22.17          & 0.8476                                 & 22.30          & 0.8552          & 0.1451          & \textbf{3.66} & 17.45          & 0.5622                                  & 17.04          & 0.6463                                 & 17.25          & 0.6043          & 0.2592          & 3.05           \\
Diff-Retinex~\cite{yi2023diff}             & 21.95          & 0.8375                                  & 21.81          & 0.8312                                 & 21.87          & 0.8337          & 0.2135          & 3.83          & 20.16          & 0.6541                                  & 19.50          & 0.6865                                 & 19.83          & 0.6753          & 0.2637          & 3.48           \\
FourierDiff~\cite{lv2024fourier}              & 20.47          & 0.7320                                  & 19.07          & 0.7145                                 & 19.63          & 0.7215          & 0.2895          & 4.14          & 17.93          & 0.5784                                  & 17.17          & 0.6106                                 & 17.55          & 0.5945          & 0.3314          & 3.36           \\
Sagiri~\cite{li2024sagiri}                   & 20.22          & 0.6915                                  & 20.07          & 0.6925                                 & 20.13          & 0.6921          & 0.2207          & 3.79          & 19.63          & 0.6175                                  & 18.23          & 0.6469                                 & 18.93          & 0.6322          & 0.2283          & 3.42           \\
PASD~\cite{PASD}                     & 21.47          & 0.6943                                  & 20.77          & 0.6800                                 & 21.05          & 0.6857          & 0.1964          & 3.88          & 18.86          & 0.5973                                  & 17.56          & 0.6353                                 & 18.21          & 0.6113          & 0.2351          & 3.39           \\
StableSR~\cite{StableSR}                 & 21.85          & 0.7028                                  & 21.67          & 0.7156                                 & 21.74          & 0.7105          & 0.1892          & 3.80          & 19.83          & 0.6139                                  & 17.63          & 0.6349                                 & 18.73          & 0.6244          & 0.2265          & 3.53           \\
DiffBIR~\cite{lin2024diffbir}                  & 21.63          & 0.7011                                  & 21.56          & 0.7063                                 & 21.59          & 0.7042          & 0.1931          & 3.76          & 19.52          & 0.5948                                  & 17.76          & 0.6366                                 & 18.64          & 0.6157          & 0.2274          & 3.42           \\
CoTF~\cite{CoTF}                     & \textbf{23.36} & 0.8630                                  & \textbf{23.49} & 0.8793                                 & \textbf{23.44} & 0.8728          & 0.1232          & 3.70          & 22.90          & 0.7029                                  & 20.13          & 0.7274                                 & 21.51          & 0.7151          & 0.1924          & 3.13           \\
Ours                     & 23.27          & 0.8654                                  & 23.31          & \textbf{0.8795}                        & 23.29          & \textbf{0.8736} & \textbf{0.1208} & 3.74          & \textbf{23.59} & \textbf{0.7106}                         & \textbf{21.56} & \textbf{0.7377}                        & \textbf{22.57} & \textbf{0.7241} & \textbf{0.1668} & 3.28           \\
\bottomrule
\end{tabular}
}
\end{table*}

\begin{table}
\centering
\caption{Quantitative Comparison on LCDP Datasets in Terms of PSNR, SSIM, LPIPS, and NIQE. The Best Results Are Highlighted in Bold.}
\label{table:lcdp}
\begin{tabular}{ccccc} 
\toprule
Methods      & PSNR           & SSIM            & LPIPS           & NIQE             \\ 
\midrule
HE~\cite{he}           & 15.98          & 0.6840          & 0.3871          & 3.7327           \\
CLAHE~\cite{clahe}         & 16.33          & 0.6420          & 0.5054          & 3.5745           \\
LIME~\cite{LIME}         & 17.34          & 0.6860          & 0.2759          & 3.4205           \\
WVM~\cite{WVM}          & 18.16          & 0.7390          & 0.2123          & 3.3705           \\
RetinexNet~\cite{retinexnet}   & 16.20          & 0.6304          & 0.2940          & 3.8659           \\
URetinexNet~\cite{uretinex}  & 17.67          & 0.7369          & 0.2504          & 3.4242           \\
DRBN~\cite{DRBN}         & 15.47          & 0.6979          & 0.3149          & 3.7554           \\
SID~\cite{SID}         & 21.89          & 0.8082          & 0.1781          & 4.2326           \\
MSEC~\cite{msec}         & 17.07          & 0.6428          & 0.3151          & 3.4499           \\
Zero-DCE~\cite{ZeroDCE}      & 18.96          & 0.7743          & 0.2055          & 3.2817           \\
Zero-DCE++~\cite{zerodce_plus}   & 18.42          & 0.7669          & 0.2204          & 3.3543           \\
RUAS~\cite{RURS}         & 13.93          & 0.6340          & 0.3458          & 3.8307           \\
SCI~\cite{sci}          & 15.96          & 0.6646          & 0.2913          & 3.4913           \\
PairLIE~\cite{pairlie}      & 16.51          & 0.6667          & 0.2945          & 3.4725           \\
ENC-SID~\cite{enc}      & 22.66          & 0.8195          & 0.1631          & 3.2771           \\
ENC-DRBN~\cite{enc}     & 23.08          & 0.8302          & 0.1536          & 3.2343           \\
CLIP-LIT~\cite{cliplit}     & 19.24          & 0.7477          & 0.2262          & 3.3858           \\
FECNet~\cite{fecnet}       & 22.34          & 0.8038          & 0.2334          & 3.6465           \\
LCDPNet~\cite{lcdpnet}      & 23.24          & 0.8420          & 0.1368          & 3.2723           \\
Diff-Retinex~\cite{yi2023diff} & 22.12          & 0.8126          & 0.1931          & 3.5782           \\
FourierDiff~\cite{lv2024fourier}  & 20.13          & 0.7356          & 0.2218          & 3.8461           \\
Sagiri~\cite{li2024sagiri}       & 21.47          & 0.6846          & 0.1845          & 3.5217           \\
PASD~\cite{PASD}         & 21.05          & 0.6736          & 0.1994          & 3.7129           \\
StableSR~\cite{StableSR}     & 21.93          & 0.7020          & 0.1749          & 3.4247           \\
DiffBIR~\cite{lin2024diffbir}      & 21.56          & 0.6831          & 0.1874          & 3.4306           \\
CoTF~\cite{CoTF}         & 23.83          & 0.8569          & 0.1035          & \textbf{3.2123}  \\
Ours         & \textbf{24.09} & \textbf{0.8627} & \textbf{0.0905} & 3.3592           \\
\bottomrule
\end{tabular}
\vspace{-3mm}
\end{table}

\subsubsection{Implementation Details}
We employed the Stable Diffusion v2-base as our foundational pre-trained model and achieved stable optimization through the two-stage training strategy. In the first stage, we configured the learning rate to \(5 \times 10^{-5}\), the batch size to 4, and the image patch size to $64 \times 64$. The Adam\cite{adam} optimizer with parameters set to \(\beta_1 = 0.9\) and \(\beta_2 = 0.999\) was utilized to update the parameters. The first stage comprised a total of 100K iterations. During the second stage, the initial learning rate was set to \(4 \times 10^{-4}\) and was gradually reduced by a cosine annealing schedule\cite{cosine}. Here, the batch size was 4, the input patch size was set to $512 \times 512$, and the total number of iterations was 300K. For the regression loss function \(L_{\text{reg}}\), the contributions of different loss terms were balanced by parameters \(\alpha = 0.1\) and \(\beta = 0.5\). The input image was randomly flipped and rotated for data augmentation.

\subsection{Comparisons with State-of-the-Art Methods}
To evaluate the effectiveness of our proposed method, we conduct a comprehensive comparative analysis of existing exposure correction techniques. Specifically, we benchmarked our approach against state-of-the-art methods, including
HE~\cite{he},
CLAHE~\cite{clahe},
LIME~\cite{LIME} (TIP'17),
WVM~\cite{WVM} (CVPR'16),
RetinexNet~\cite{retinexnet} (BMVC'18),
URetinexNet~\cite{uretinex} (CVPR'22),
DRBN~\cite{DRBN} (CVPR'20),
SID~\cite{SID} (CVPR'18),
MSEC~\cite{msec} (CVPR'21),
Zero-DCE~\cite{ZeroDCE} (CVPR'20),
Zero-DCE++~\cite{zerodce_plus} (TPAMI'21),
RUAS~\cite{RURS} (CVPR'21),
SCI~\cite{sci} (CVPR'22),
PairLIE~\cite{pairlie} (CVPR'23),
ENC~\cite{enc} (CVPR'22),
CLIP-LIT~\cite{cliplit} (ICCV'23),
FECNet~\cite{fecnet} (ECCV'22),
LCDPNet~\cite{lcdpnet} (ECCV'22),
Diff-Retinex~\cite{yi2023diff} (ICCV'23),
FourierDiff~\cite{lv2024fourier}(CVPR'24),
Sagiri~\cite{li2024sagiri} (arxiv'24),
PASD~\cite{PASD}(ECCV'24),
StableSR~\cite{StableSR} (IJCV'24),
DiffBIR~\cite{lin2024diffbir} (ECCV'24)
and
CoTF~\cite{CoTF} (CVPR'24).
We employ the Peak Signal-to-Noise Ratio (PSNR) to evaluate the image fidelity, the Structural Similarity Index (SSIM)\cite{ssim} for measuring the structural similarity of images, the Learned Perceptual Image Patch Similarity (LPIPS)\cite{lpips} to assess the perceptual quality of the images, and the Naturalness Image Quality Evaluator (NIQE)\cite{niqe} to measure the naturalness of images in a no-reference manner.

\begin{table}
\centering
\caption{Quantitative comparison with state-of-the-art methods on three unsupervised datasets (LIME, DICM, NPE), with NIQE as the evaluation metric (lower values indicate better performance).}
\label{table:unlabeled}
\begin{tabular}{cccc} 
\toprule
Methods & LIME~         & DICM          & NPE            \\ 
\midrule
HE~\cite{he}           & 4.01          & 3.64          & 3.72           \\
CLAHE~\cite{clahe}        & 3.91          & 3.62          & 3.78           \\
LIME~\cite{LIME}         & 4.16          & 3.85          & 4.25           \\
WVM~\cite{WVM}          & 3.79          & 3.90          & 4.16           \\
RetinexNet~\cite{retinexnet}   & 4.42          & 4.20          & 4.59           \\
URetinexNet~\cite{uretinex}  & 3.51          & 3.57          & 3.72           \\
DRBN~\cite{DRBN}         & 3.72          & 3.64          & 3.82           \\
SID~\cite{SID}          & 4.14          & 3.72          & 3.68           \\
MSEC~\cite{msec}         & 3.76          & 3.68          & 3.49           \\
Zero-DCE~\cite{ZeroDCE}      & 3.49          & 3.29          & 3.74           \\
Zero-DCE++~\cite{zerodce_plus}   & 3.93          & 3.46          & 3.97           \\
RUAS~\cite{RURS}         & 4.07          & 5.93          & 5.88           \\
SCI~\cite{sci}          & 3.92          & 3.73          & 3.89           \\
PairLIE~\cite{pairlie}      & 3.81          & 4.18          & 4.30           \\
ENC-SID~\cite{enc}      & 3.34          & 3.51          & 4.12           \\
ENC-DRBN~\cite{enc}     & 3.28          & 3.49          & 4.31           \\
CLIP-LIT~\cite{cliplit}     & 3.45          & 3.72          & 3.85           \\
FECNet~\cite{fecnet}       & 3.61          & 3.43          & 3.78           \\
LCDPNet~\cite{lcdpnet}      & 3.31          & \textbf{3.02} & 3.47           \\
Diff-Retinex~\cite{yi2023diff} & 3.42          & 3.37          & 3.56           \\
FourierDiff~\cite{lv2024fourier}  & 3.79          & 3.72          & 3.83           \\
Sagiri~\cite{li2024sagiri}       & 3.31          & 3.26          & 3.52           \\
PASD~\cite{PASD}         & 3.27          & 3.42          & 3.61           \\
StableSR~\cite{StableSR}     & 3.30          & 3.35          & 3.48           \\
DiffBIR~\cite{lin2024diffbir}      & 3.22          & 3.28          & 3.49           \\
CoTF~\cite{CoTF}         & \textbf{3.17} & 3.10          & 3.53           \\
Ours         & 3.25          & 3.22          & \textbf{3.37}  \\
\bottomrule
\end{tabular}
\vspace{-3mm}
\end{table}

\subsubsection{Quantitative Evaluation}

\begin{figure*}[t]
    \centering
    \includegraphics[width = 1.0\textwidth]{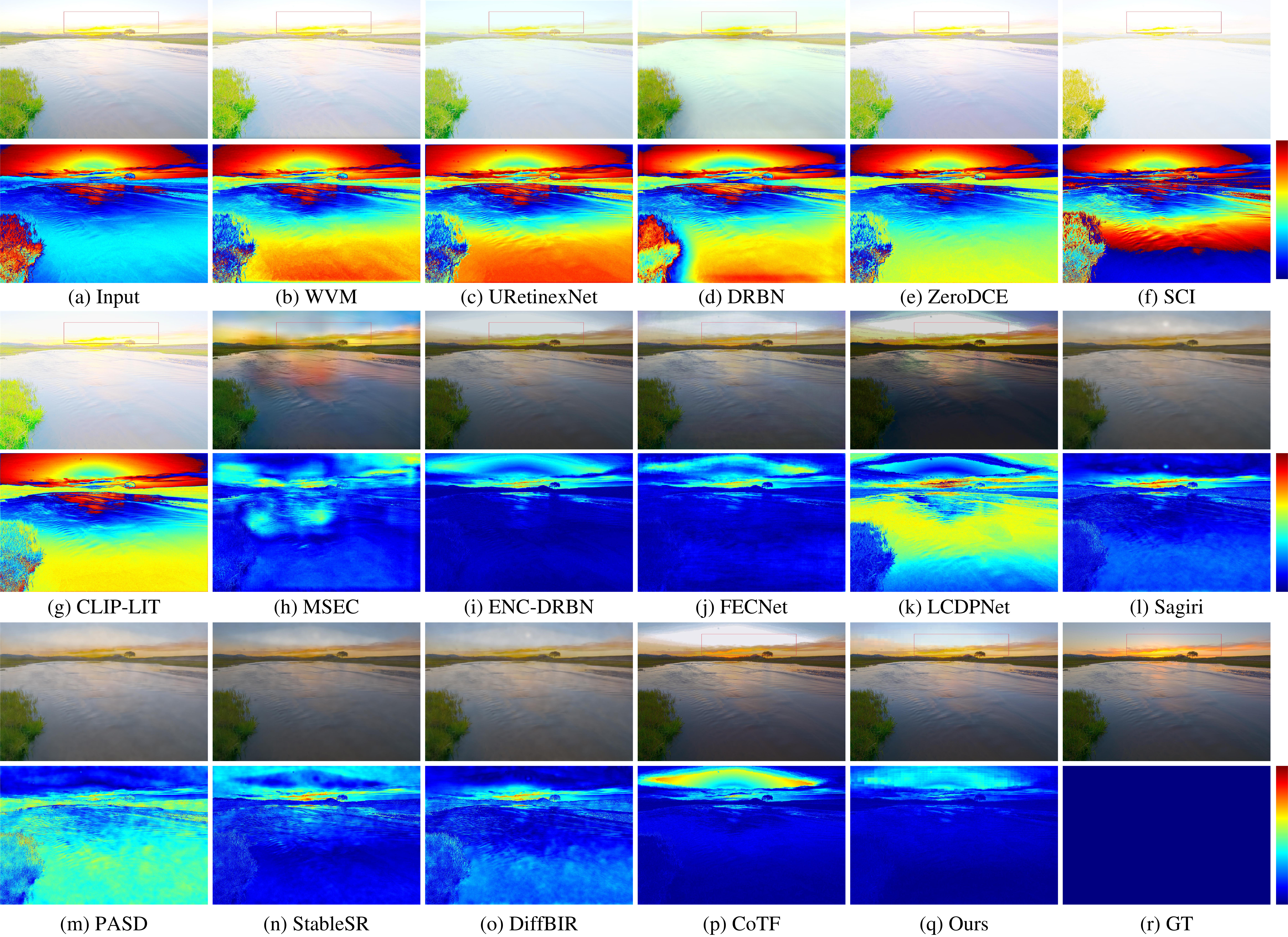}
    \caption{Visual comparison with state-of-the-art methods on overexposed images from the SICE dataset. Best viewed by zooming in.}
    \label{fig:sice1}
    \vspace{-3mm}
\end{figure*}

\begin{figure*}[t]
    \centering
    \includegraphics[width = 1.0\textwidth]{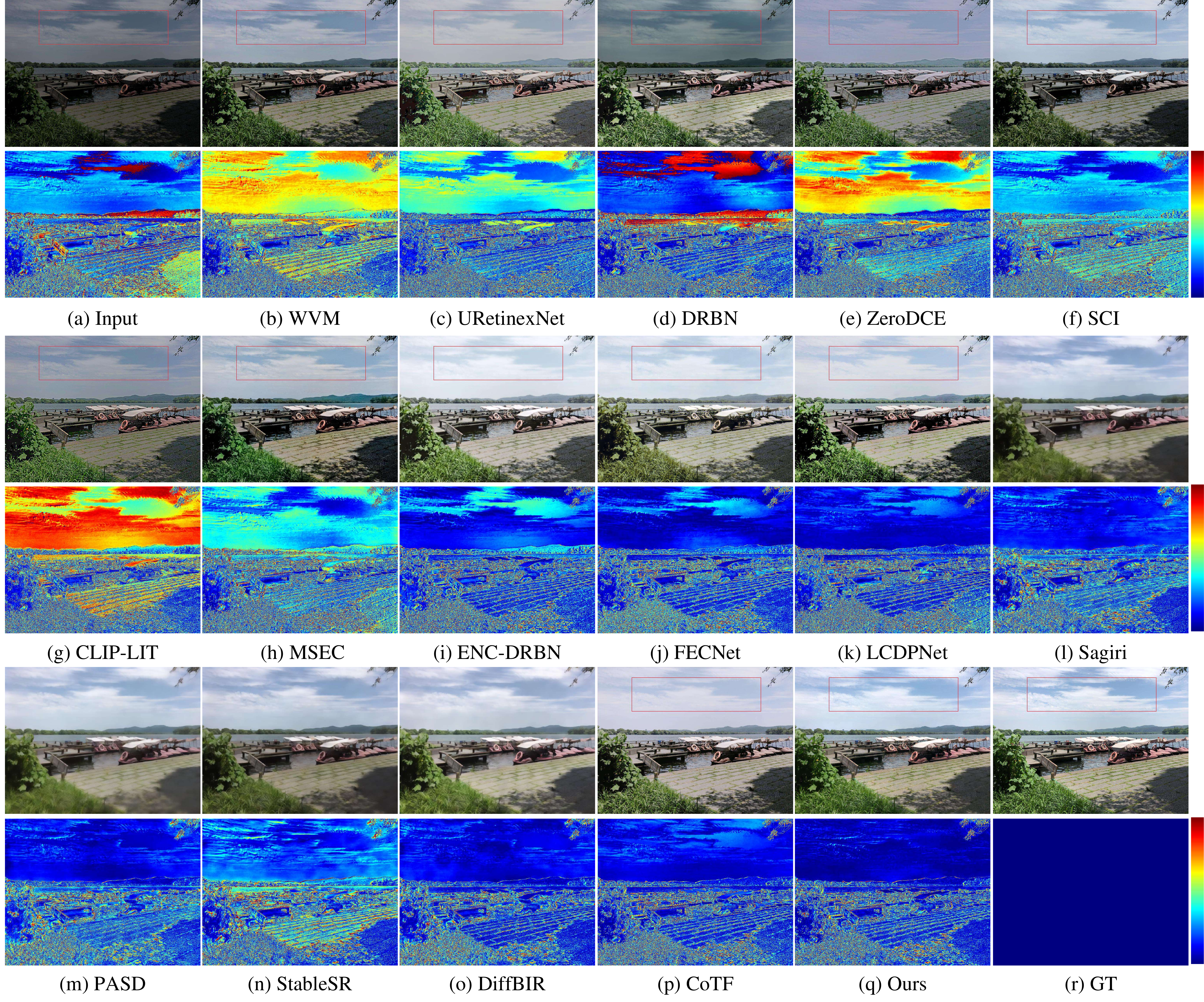}
    \caption{Visual comparison with state-of-the-art methods on underexposed images from the SICE dataset. Best viewed by zooming in.}
    \label{fig:sice2}
    \vspace{-3mm}
\end{figure*}

Table~\ref{table:msec_sice} provides a quantitative comparison of our method against state-of-the-art methods on the MSEC and SICE datasets. The top-performing metrics are highlighted in bold for clarity. It is evident that our method surpasses previous approaches in overall performance for multi-exposure correction, especially in terms of perceptual quality. For the MSEC dataset, we averaged the first two exposure levels to form the underexposed subset and the remaining exposure levels as the overexposed subset. Within both the overexposed and underexposed subsets, our method achieves the highest or near-highest PSNR, SSIM and LPIPS scores. Across the entire MSEC dataset, our method demonstrates the best perceptual quality as indicated by the LPIPS score of 0.1208, while also maintaining competitive PSNR value of 23.29 dB and SSIM value of 0.8736. Similarly, for the SICE dataset, our method yields the superior average performance, attaining an average PSNR value of 22.57 dB, an SSIM value of 0.7241, and an LPIPS value of 0.1668. 
Table~\ref{table:lcdp} presents a quantitative comparison of various methods on the LCDP dataset. The results clearly indicate that our proposed method achieves the best performance in correcting non-uniform exposures, as reflected by the highest scores across all evaluation metrics. Notably, our method outperforms the second-best approach with a 0.2 dB improvement in PSNR. 
Table~\ref{table:unlabeled} presents the generalization performance of various methods on the LIME, DICM, and NPE datasets. It can be observed that our proposed method also demonstrates strong competitiveness in terms of generalization ability.
The results consistently demonstrate the effectiveness of our approach, which integrates regression models and diffusion models with robust diffusion generation priors, achieves superior performance in exposure correction across diverse lighting conditions.

\subsubsection{Qualitative Evaluation}

\begin{figure*}[htbp]
    \centering
    \includegraphics[width = 1.0\textwidth]{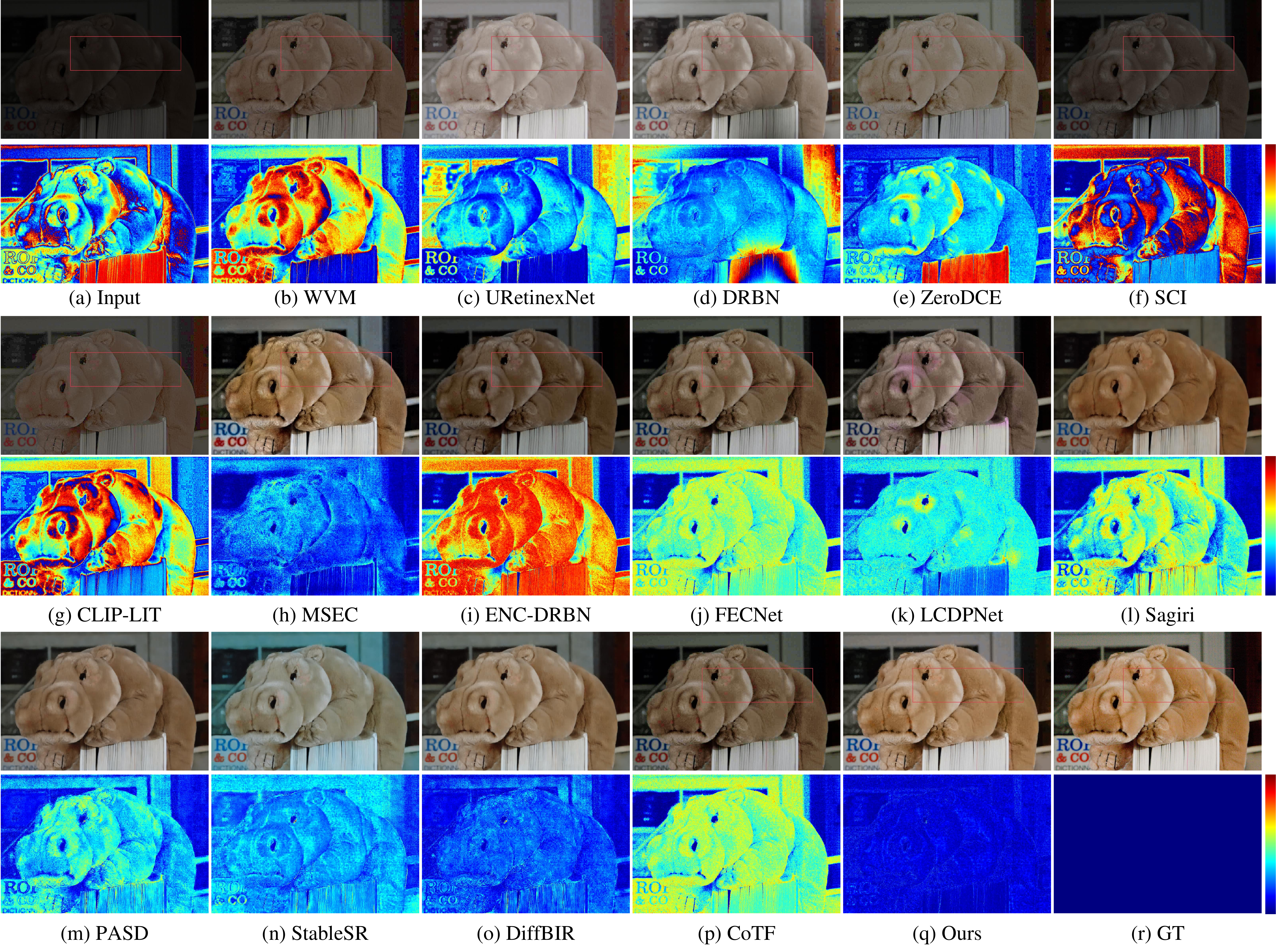}
    \caption{Visual comparison with state-of-the-art methods on overexposed images from the MSEC dataset. Best viewed by zooming in.}
    \label{fig:msec1}
    \vspace{-3mm}
\end{figure*}

\begin{figure}[htbp]
    \centering
    \includegraphics[width = 0.5\textwidth]{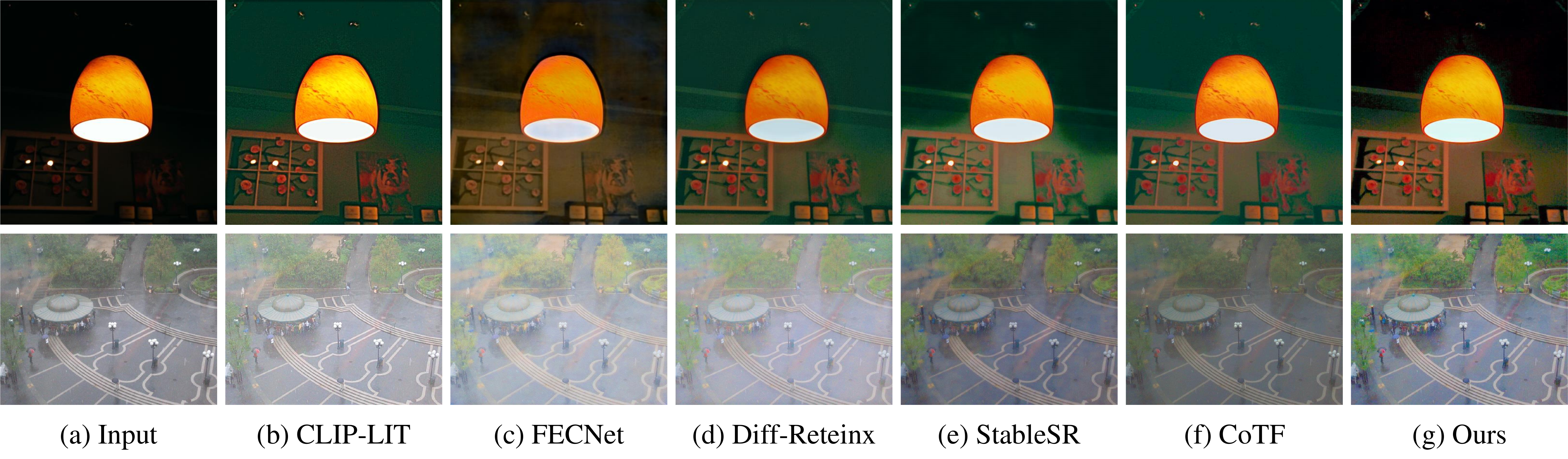}
    \caption{Visual comparison with state-of-the-art methods on the unsupervised dataset
}
    \label{fig:unlabeled}
    \vspace{-4mm}
\end{figure}

\begin{figure}[htbp]
    \centering
    \includegraphics[width = 0.5\textwidth]{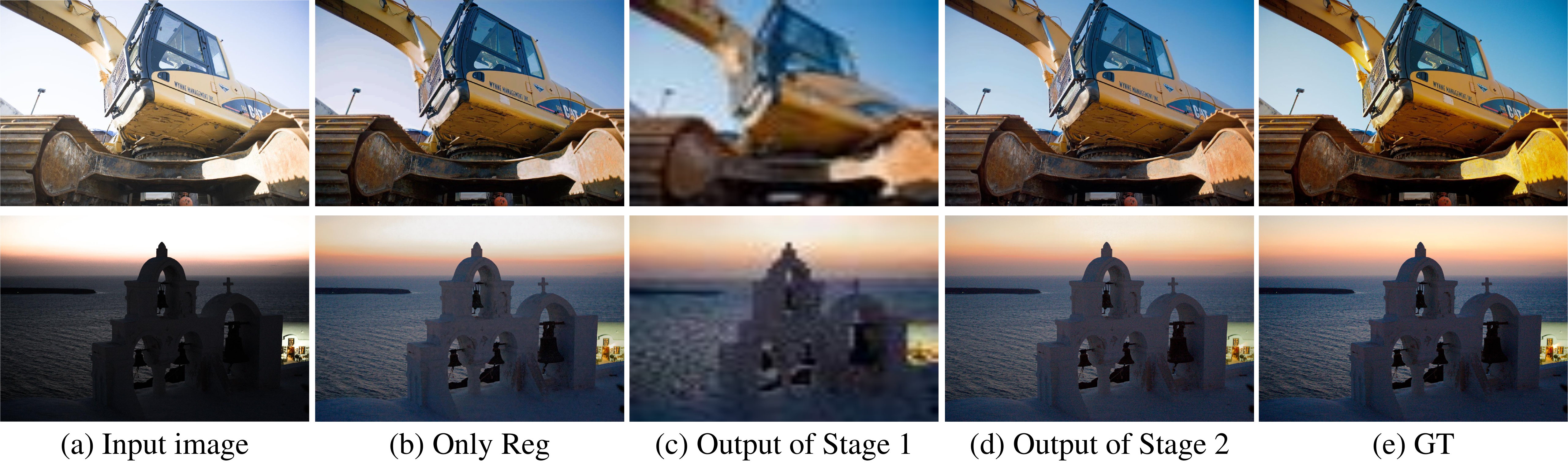}
    \caption{
Ablation analysis on the diffusion and regression models. The output of the diffusion model in the first stage is a low-resolution $64 \times 64$ image. Compared to a pure regression model, it can recover some content in regions with extreme exposure. We also feed diffusion prior features with richer knowledge into the second stage.
}
    \label{fig:abla}
    \vspace{-4mm}
\end{figure}

\begin{figure*}[htbp]
    \centering
    \includegraphics[width = 1.0\textwidth]{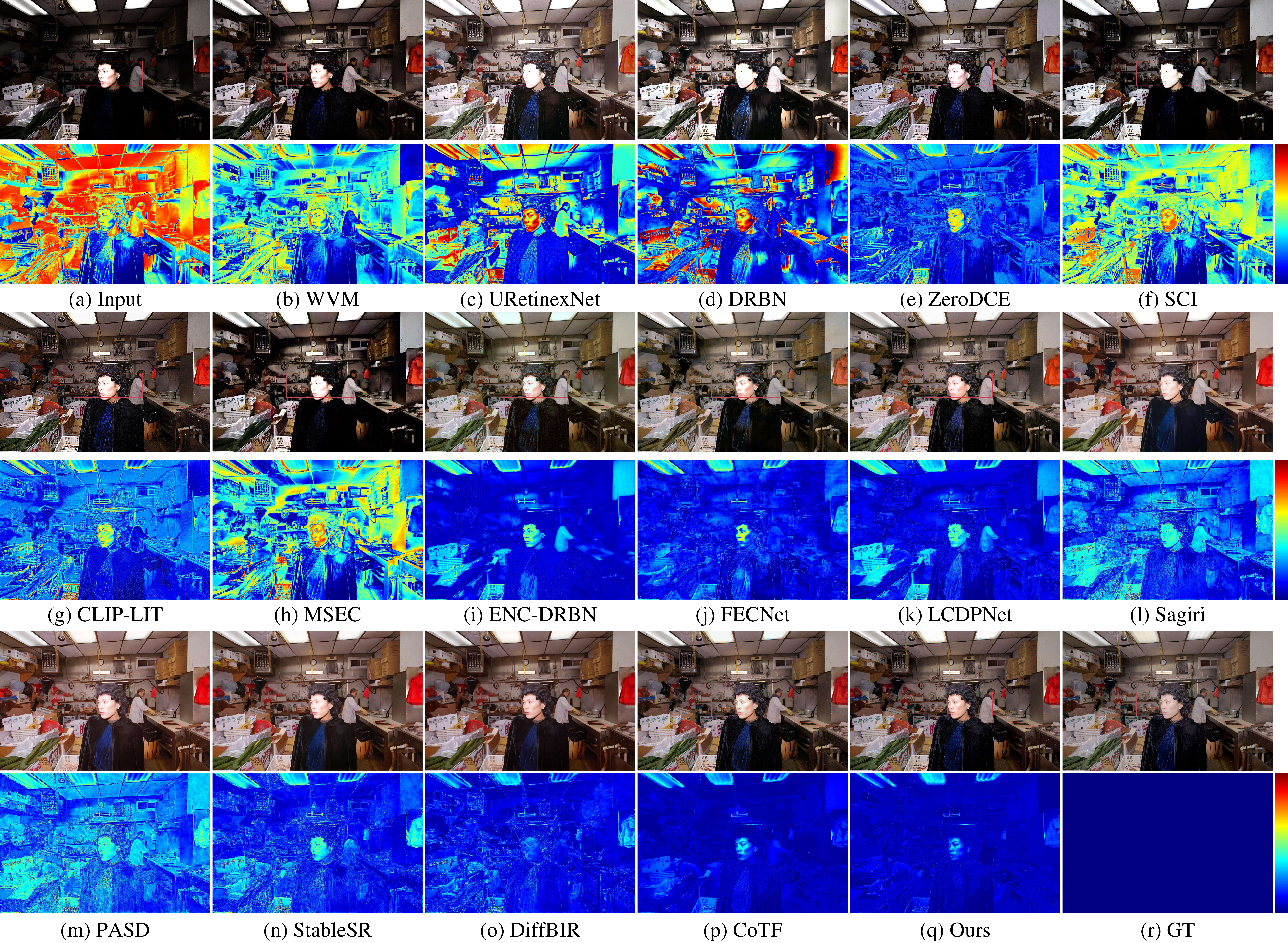}
    \caption{Visual comparison with state-of-the-art methods on non-uniform illumination images from the LCDP dataset. Best viewed by zooming in.}
    \label{fig:lcdp1}
    \vspace{-3mm}
\end{figure*}

\begin{figure}[htbp]
    \centering
    \includegraphics[width = 0.5\textwidth]{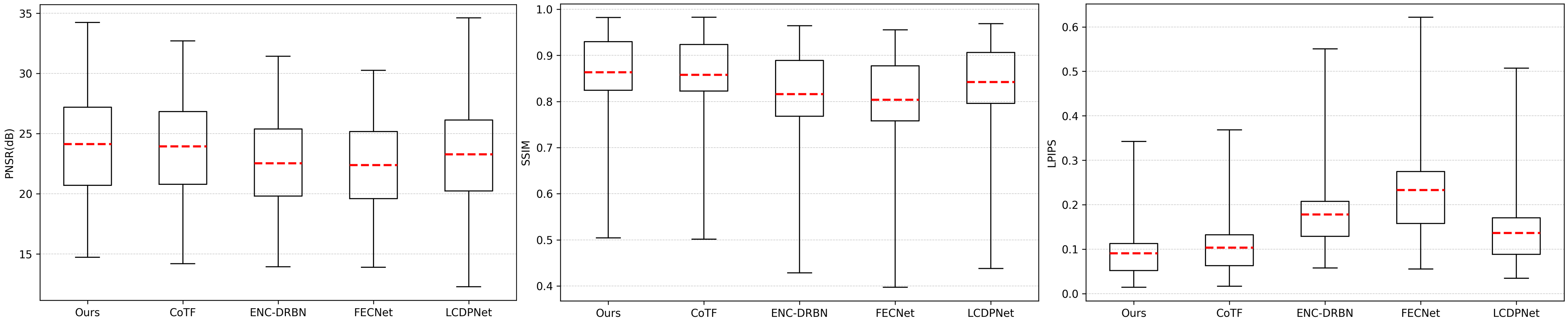}
    \caption{Boxplots of PSNR, SSIM and LPIPS on the LCDP dataset compared with state-of-the-art methods.
}
    \label{fig:boxplot}
    \vspace{-4mm}
\end{figure}

\begin{figure*}[htbp]
    \centering
    \includegraphics[width = 1.0\textwidth]{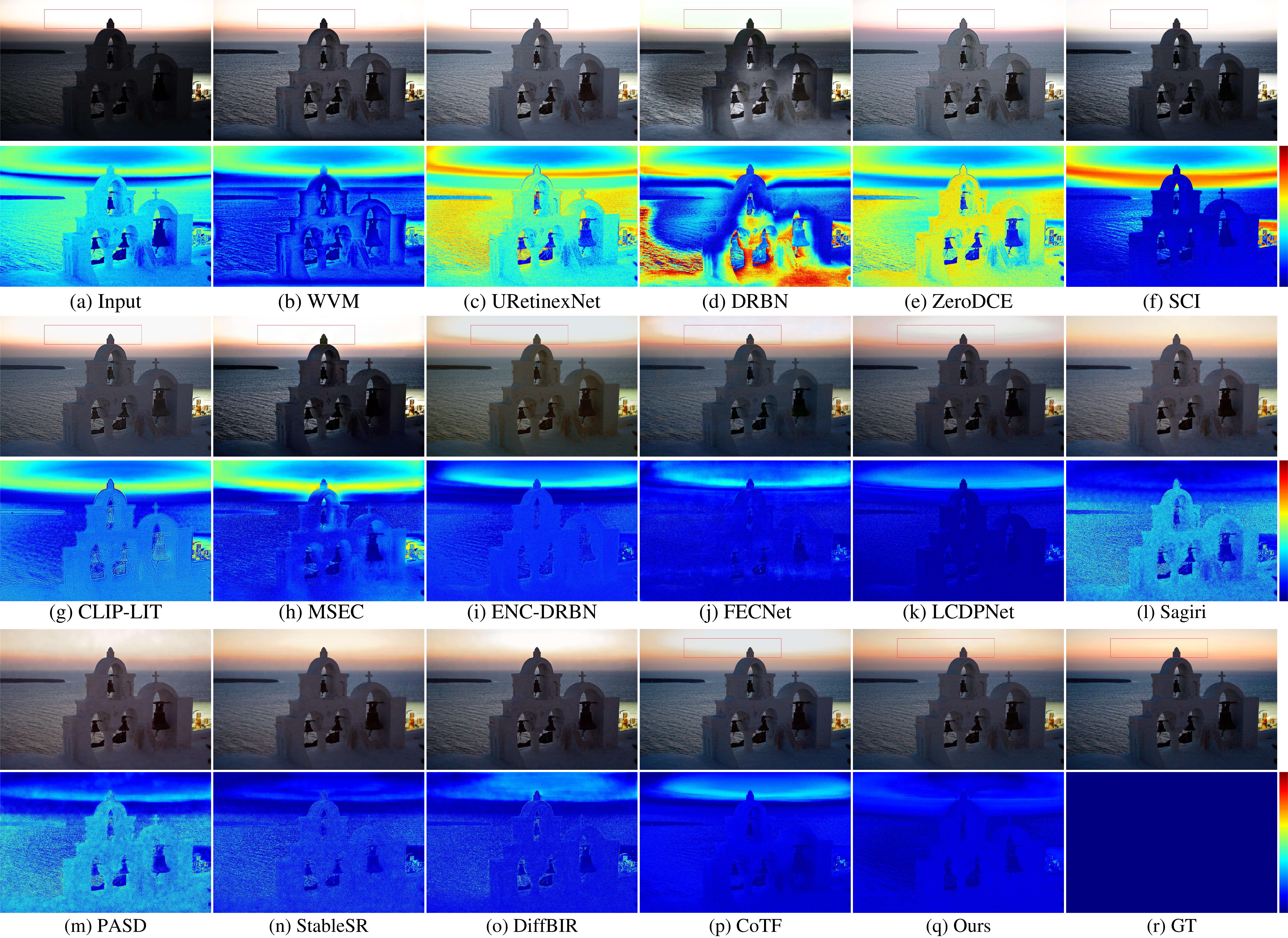}
    \caption{Visual comparison with state-of-the-art methods on non-uniform illumination images from the LCDP dataset. Best viewed by zooming in.}
    \label{fig:lcdp2}
    \vspace{-3mm}
\end{figure*}

We provide comprehensive visual comparisons of our method against state-of-the-art methods on several datasets, showcasing qualitative results. We employ heatmaps to provide a clearer visualization of performance differences.
Fig.~\ref{fig:sice1} illustrates the visual effects of different methods on overexposed images from the SICE dataset. It can be observed that some methods (e.g., the first row) failed to effectively mitigate overexposure issues, resulting in the loss of image details. While the MSEC achieves some improvement in exposure, its output exhibited noticeable color distortions, compromising the realism and visual consistency. Additionally, methods such as ENC-DRBN, FECNet, LCDPNet, and CoTF make brightness adjustments but fail to restore details in sky regions, leaving large blank areas that detract from overall visual quality. In contrast, our method consistently enhances the image quality and restores smooth transitions in the sky region.
Fig.~\ref{fig:sice2} shows the results of different methods for underexposed images from the SICE dataset. Most existing methods suffer from issues of either over-enhancement or under-enhancement, leading to unnatural results. In contrast, our method effectively avoids these issues, producing images with balanced brightness adjustment that appear natural and visually pleasing.

Fig.~\ref{fig:msec1} shows the visual effects of different methods on the MSEC dataset. For underexposed images, the majority of methods suffer from noticeable color casts, whereas our method demonstrates superior color restoration with appropriate brightness.
In the LCDP dataset, the visual effects of non-uniform illumination correction are shown in Fig.~\ref{fig:lcdp1} and Fig.~\ref{fig:lcdp2}. In Fig.~\ref{fig:lcdp1}, focusing the facial region, it is evident that our method effectively avoids saturation issues caused by intense lighting, preserving rich texture details. In the sky region of Fig.~\ref{fig:lcdp2}, our method successfully restores the vibrant colors of the sunset, adding depth and realism to the scene. Overall, our method demonstrates remarkable adaptability in handling overexposed highlights and underexposed shadows, producing natural and pleasing visual effects.
Fig.~\ref{fig:unlabeled} shows the visual results on unlabeled datasets. The results indicate that our method generates images with superior visual quality, color fidelity, and contrast.

These results demonstrate that our method, which combines generative priors with regression techniques, exhibits outstanding performance in high-quality image exposure correction.

\subsection{Ablation Study}
To validate the effectiveness of our design, we conduct ablation experiments on components of the proposed method on the LCDP dataset.

\subsubsection{The effectiveness of fine-tuning strategy}
The proposed fine-tuning strategy achieves single-step image generation with conditional control for degraded images by applying architectural modifications and focusing on denoising tasks at specific noise levels. To verify the effectiveness and generality of the proposed fine-tuning strategy, this experiment adopts the same configuration as existing fine-tuning methods. While retaining the VAE component, we removed the regression model to independently assess the actual contribution of the fine-tuning strategy. Quantitative results are presented in Table~\ref{table:fine_tuning}. Specifically, compared to single-step sampling with multi-noise training methods, our method demonstrates better performance in single-step generation. This improvement stems from the fact that the model only needs to handle one specific noise intensity during training, rather than multiple noise intensities, which simplifies the learning task and training difficulty. In addition, our single-step generation not only achieves performance comparable to original 20-step DDIM, but also significantly reduces the inference time and improves the sampling efficiency. This efficiency improvement makes our solution more competitive and practical for real-world applications.

\begin{table}
    \centering
    \caption{Ablation study on the effectiveness of the fine-tuning strategy}
    \label{table:fine_tuning}
    \begin{tabular}{c!{\vrule width \lightrulewidth}c!{\vrule width \lightrulewidth}ccc}
        \toprule
        Train setting & Sampling steps & PSNR $\uparrow$  & SSIM $\uparrow$   & LPIPS $\downarrow$           \\
        \midrule
        Multi-noise   & 20              & 22.03          & 0.6654          & 0.1740          \\
        Multi-noise   & 1               & 21.58          & 0.6577          & 0.1782          \\
        Single-noise  & 1               & \textbf{21.97} & \textbf{0.6652} & \textbf{0.1757} \\
        \bottomrule
    \end{tabular}
    \vspace{-3mm}
\end{table}

\subsubsection{The effect of the noise level}
Considering the need to set a fixed noise level, we conducted an ablation study to investigate the effect of different noise levels. In this ablation experiment, we compared the results of adding noise over different time steps, including 1000, 500 and 1. As shown in Table~\ref{table:noise_level}, the best results are achieved when the noise addition timestep is set to 1000 steps. This is primarily because the test input is pure Gaussian noise, which closely resembles the noise distribution under the 1000-steps condition. Aligning the noise addition strategy with the input characteristics of the test environment helps to maintain the model's performance. In contrast, lower noise levels result in a significant performance drop. Therefore, we selected the 1000-steps noise level as the denoising target.

\begin{table}
    \centering
    \caption{Ablation study on the effect of the noise level.}
    \label{table:noise_level}
    \begin{tabular}{c!{\vrule width \lightrulewidth}ccc}
        \toprule
        $t$    & PSNR $\uparrow$           & SSIM $\uparrow$           & LPIPS $\downarrow$         \\
        \midrule
        1000 & \textbf{21.97} & \textbf{0.6652} & \textbf{0.1757} \\
        500  & 19.45          & 0.5932          & 0.2452            \\
        1    & 6.48           & 0.0650           & 0.5739            \\
        \bottomrule
    \end{tabular}
    \vspace{-4mm}
\end{table}

\begin{table}
    \centering
    \caption{Ablation study on the effectiveness of integrating diffusion models with regression methods.}
    \label{table:diffusion_regression}
    \begin{tabular}{c!{\vrule width \lightrulewidth}ccc}
        \toprule
        Setting        & PSNR $\uparrow$           & SSIM $\uparrow$           & LPIPS $\downarrow$ \\
        \midrule
        Only SD (Down) & 21.18          & 0.6487          & 0.1849          \\
        Only SD (VAE)  & 21.97          & 0.6652          & 0.1757          \\
        Only Regression & 23.87         & 0.8579          & 0.1043          \\
        Single Scale   & 24.01          & 0.8615          & 0.0933          \\
        Ours           & \textbf{24.09} & \textbf{0.8627} & \textbf{0.0905} \\
        \bottomrule
    \end{tabular}
    \vspace{-4mm}
\end{table}

\begin{table*}
\centering
\caption{Comparison of inference times for different methods. All methods were tested on $512 \times 512$ resolution images, with each method run 10 times and the results averaged.}
\resizebox*{\linewidth}{!}{
\begin{tabular}{cccccccccccc} 
\toprule
Method  & RetinexNet & URetinexNet & DRBN    & SID          & MSEC        & Zero-DCE & RUAS   & SCI      & PairLIE & ENC-SID & ENC-DRBN  \\
Time(s) & 0.1529     & 0.1877      & 0.1226  & 0.0387       & 0.0468      & 0.0229  & 0.0281 & 0.0021   & 0.0716  & 0.0647  & 0.1869    \\ 
\midrule
Method  & CLIP-LIT   & FECNet      & LCDPNet & Diff-Retinex & FourierDiff & Sagiri  & PASD   & StableSR & DiffBIR & CoTF    & Ours      \\
Time(s) & 0.0877     & 0.1261      & 0.0472  & 24.3572      & 26.2911     & 9.5386  & 9.4814 & 25.1425  & 12.6973 & 0.0071  & 0.3096    \\
\bottomrule
\end{tabular}}
\label{table:time}
\end{table*}

\subsubsection{The effectiveness of integrating diffusion models with regression models}
In this section, we perform ablation studies to evaluate the effectiveness of integrating diffusion models with regression models. Table~\ref{table:diffusion_regression} presents the quantitative results. First, we fine-tune Stable Diffusion independently, without introducing a regression model. Due to memory constraints, we downsample the input image before passing it to the denoising UNet. 
In the ``Only SD (Down)" setting, we use bilinear downsampling to reduce the input image resolution, perform the diffusion process at a lower resolution, and then upsample the results. In the ``Only SD (VAE)" setting, we use the VAE encoder and decoder for image compression and reconstruction.
Second, we evaluate the performance of the regression model alone for exposure correction, labeled as ``Only Regression", which relies solely on regression loss. The ``Single Scale" setting refers to the integration of diffusion prior features in a single-scale manner. Finally, we combine the diffusion prior feature with the regression model in a multi-scale fashion, adopting a two-stage training strategy. 

By comparing the first and second rows in the table, we observe that, compared to conventional downsampling methods, VAE slightly retains more information during the compression and reconstruction process. However, VAE still has significant room for improvement in fidelity and introduces additional computational overhead. As a result, we chose to remove VAE from our method. The regression-only model performs well in terms of fidelity but still has room for improvement in perceptual quality. Furthermore, the single-scale integration strategy performs worse than the multi-scale strategy, indicating that multi-scale diffusion prior features can better restore degraded images. Our complete model demonstrates outstanding performance in both fidelity and perceptual quality. Fig.~\ref{fig:abla} presents the visualization results of this ablation study. It can be seen that after incorporating the diffusion prior features, the model is able to generate outputs with higher perceptual quality. This result shows that our method successfully integrates the strengths of diffusion models and regression models, thereby improving the overall performance.

\subsubsection{The effectiveness of joint cross-attention module}
In this section, we conduct an ablation study to verify the effectiveness of the Joint Cross Attention Module (JCAM). Specifically, we compare JCAM with the previously proposed Cross Attention Module (CAM) \cite{CoTF}, which employs one feature as the query (Q) and the other feature as the key(K) and value (V). The experimental setup includes two modes: $\text{CAM}_{diff}$, where diffusion features are used as the query, and $\text{CAM}_{reg}$, where regression features are used as the query. As shown in Table \ref{table:cross_attention}, the proposed JCAM consistently outperforms CAM across multiple performance metrics. These results demonstrate that our method is more effective in integrating diffusion features and regression features, thereby improving overall performance.

\begin{table}
    \centering
    \caption{Ablation study on the effectiveness of joint cross-attention module.}
    \label{table:cross_attention}
    \begin{tabular}{c!{\vrule width \lightrulewidth}ccc}
        \toprule
        Setting         & PSNR $\uparrow$  & SSIM $\uparrow$   & LPIPS $\downarrow$           \\
        \midrule
        $\text{CAM}_{reg}$  & 23.59          & 0.8510          & 0.1113         \\
        $\text{CAM}_{diff}$  & 23.96         & 0.8602          & 0.0982         \\
        JCAM       & \textbf{24.09} & \textbf{0.8627} & \textbf{0.0905} \\
        \bottomrule
    \end{tabular}
    \vspace{-4mm}
\end{table}

\subsubsection{Performance Distribution Analysis}
To further validate the superiority of our method, we present boxplots to illustrate the performance distribution, as shown in Fig.~\ref{fig:boxplot}. It can be observed that our method not only achieves higher average performance but also demonstrates advantages in the overall performance distribution.

\subsubsection{Inference time}
We also report the inference times of different methods for image processing (see Table~\ref{table:time}). For images with a resolution of $512 \times 512$, our method requires only 0.3096 seconds, demonstrating its advantage in inference speed. Compared with other diffusion-based exposure correction methods, our approach significantly reduces inference time and computational overhead by maintaining single-step diffusion inference. Moreover, even the best-performing regression methods struggle to effectively restore these challenging samples due to the lack of rich pre-trained diffusion generative priors. Consequently, our method improves perceptual quality while achieving high efficiency with single-step inference, striking a better balance between performance and efficiency relative to existing approaches.

\begin{figure}[htbp]
    \centering
    \includegraphics[width = \linewidth]{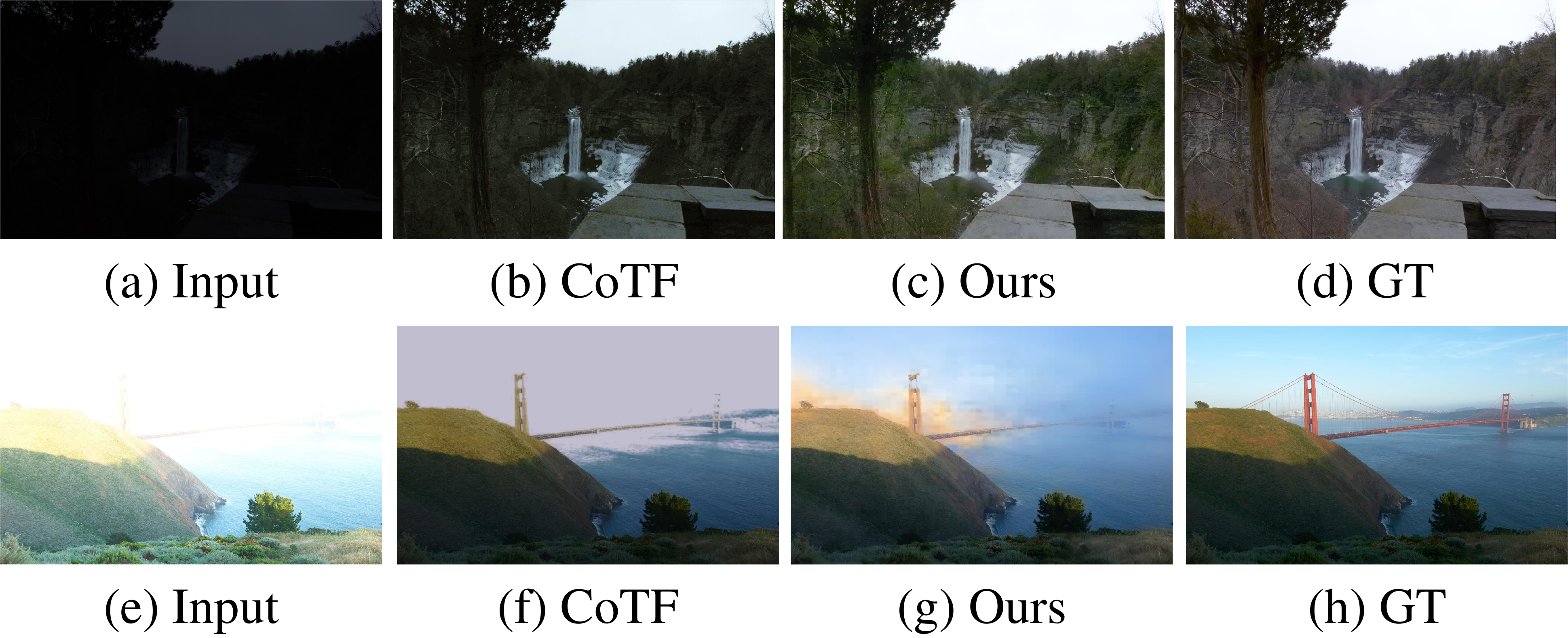}
    \caption{Failure Cases. Our approach fails to correct large areas of overexposure and underexposure, such as the color deviation in (c) and the lost details in (g).}
    \label{fig:failure}
    \vspace{-4mm}
\end{figure}

\subsection{Limitations}
Fig.~\ref{fig:failure} presents two failure cases of our method. In instances of significant overexposure or underexposure across large areas of the image, while our generative approach outperforms previous state-of-the-art methods; however, certain limitations remain. For instance, Fig.~\ref{fig:failure}(c) exhibits color deviations, and Fig.~\ref{fig:failure}(g) struggles to recover the original missing details. To address this issue, future work could explore incorporating language-guided generation models, leveraging multimodal information to generate semantically accurate structural information in extreme regions. Additionally, building a large-scale exposure correction dataset to further train the model could be beneficial. We plan to explore these directions in future work.

Furthermore, directly adopting pre-trained models such as Stable Diffusion entails substantial demands on memory and computational resource. In future research, we plan to explore more efficient solutions by developing a customized knowledge distillation framework. This framework will focus on extracting and optimizing essential feature priors from large models, enabling a significant reduction in computational and storage requirements while preserving high performance. Moreover, we are committed to translating these research outcomes into applicable technologies to drive advancements in related fields.

\section{Conclusion}
In this paper, we propose a novel Diffusion Prior-based Exposure Correction (DPEC) method, which leverages the image generation priors encapsulated in large-scale image diffusion models to achieve high-quality exposure correction. Specifically, we propose an efficient fine-tuning strategy designed to effectively derive an exposure correction model from pre-trained image diffusion models. Furthermore, we design a hierarchical integration framework that seamlessly combines diffusion models with regression models, effectively preserving image details while minimizing the risk of artifact generation. Extensive experiments, both qualitative and quantitative, demonstrate the superiority of our proposed method compared to other state-of-the-art approaches.




\bibliographystyle{IEEEtran}
\bibliography{IEEEabrv,mybibfile}

\begin{thebibliography}{10}
\providecommand{\url}[1]{#1}
\csname url@samestyle\endcsname
\providecommand{\newblock}{\relax}
\providecommand{\bibinfo}[2]{#2}
\providecommand{\BIBentrySTDinterwordspacing}{\spaceskip=0pt\relax}
\providecommand{\BIBentryALTinterwordstretchfactor}{4}
\providecommand{\BIBentryALTinterwordspacing}{\spaceskip=\fontdimen2\font plus
\BIBentryALTinterwordstretchfactor\fontdimen3\font minus
  \fontdimen4\font\relax}
\providecommand{\BIBforeignlanguage}[2]{{%
\expandafter\ifx\csname l@#1\endcsname\relax
\typeout{** WARNING: IEEEtran.bst: No hyphenation pattern has been}%
\typeout{** loaded for the language `#1'. Using the pattern for}%
\typeout{** the default language instead.}%
\else
\language=\csname l@#1\endcsname
\fi
#2}}
\providecommand{\BIBdecl}{\relax}
\BIBdecl

\bibitem{hu2023_uniad}
Y.~Hu, J.~Yang, L.~Chen, K.~Li, C.~Sima, X.~Zhu, S.~Chai, S.~Du, T.~Lin,
  W.~Wang, L.~Lu, X.~Jia, Q.~Liu, J.~Dai, Y.~Qiao, and H.~Li,
  ``Planning-oriented autonomous driving,'' in \emph{Proceedings of the
  IEEE/CVF Conference on Computer Vision and Pattern Recognition}, 2023.

\bibitem{cao2023iterative}
M.~Cao, F.~Wei, C.~Xu, X.~Geng, L.~Chen, C.~Zhang, Y.~Zou, T.~Shen, and
  D.~Jiang, ``Iterative proposal refinement for weakly-supervised video
  grounding,'' in \emph{Proceedings of the IEEE/CVF Conference on Computer
  Vision and Pattern Recognition}, 2023, pp. 6524--6534.

\bibitem{msec}
M.~Afifi, K.~G. Derpanis, B.~Ommer, and M.~S. Brown, ``Learning multi-scale
  photo exposure correction,'' in \emph{Proceedings of the IEEE Conference on
  Computer Vision and Pattern Recognition}, 2021, pp. 9157--9167.

\bibitem{lcdpnet}
H.~Wang, K.~Xu, and R.~W. Lau, ``Local color distributions prior for image
  enhancement,'' in \emph{Proceedings of the European Conference on Computer
  Vision}.\hskip 1em plus 0.5em minus 0.4em\relax Springer, 2022, pp. 343--359.

\bibitem{sice}
J.~Cai, S.~Gu, and L.~Zhang, ``Learning a deep single image contrast enhancer
  from multi-exposure images,'' \emph{IEEE Transactions on Image Processing},
  vol.~27, no.~4, pp. 2049--2062, 2018.

\bibitem{StableSR}
J.~Wang, Z.~Yue, S.~Zhou, K.~C. Chan, and C.~C. Loy, ``Exploiting diffusion
  prior for real-world image super-resolution,'' \emph{International Journal of
  Computer Vision}, 2024.

\bibitem{PASD}
T.~Yang, R.~Wu, P.~Ren, X.~Xie, and L.~Zhang, ``Pixel-aware stable diffusion
  for realistic image super-resolution and personalized stylization,''
  \emph{arXiv preprint arXiv:2308.14469}, 2024.

\bibitem{he}
S.~M. Pizer, E.~P. Amburn, J.~D. Austin, R.~Cromartie, A.~Geselowitz, T.~Greer,
  B.~ter Haar~Romeny, J.~B. Zimmerman, and K.~Zuiderveld, ``Adaptive histogram
  equalization and its variations,'' \emph{Computer vision, graphics, and image
  processing}, 1987.

\bibitem{clahe}
K.~Zuiderveld, ``Contrast limited adaptive histogram equalization,''
  \emph{Graphics gems}, 1994.

\bibitem{generalizations_HE}
J.~Stark, ``Adaptive image contrast enhancement using generalizations of
  histogram equalization,'' \emph{IEEE Transactions on Image Processing},
  vol.~9, no.~5, pp. 889--896, 2000.

\bibitem{bennett2005video}
E.~P. Bennett and L.~McMillan, ``Video enhancement using per-pixel virtual
  exposures,'' \emph{ACM Transactions on Graphics (TOG)}, vol.~24, no.~3, pp.
  845--852, 2005.

\bibitem{yuan2012automatic}
L.~Yuan and J.~Sun, ``Automatic exposure correction of consumer photographs,''
  in \emph{Proceedings of the European Conference on Computer Vision}.\hskip
  1em plus 0.5em minus 0.4em\relax Springer, 2012, pp. 771--785.

\bibitem{SSR}
D.~Jobson, Z.~Rahman, and G.~Woodell, ``Properties and performance of a
  center/surround retinex,'' \emph{IEEE Transactions on Image Processing},
  vol.~6, no.~3, pp. 451--462, 1997.

\bibitem{NPE}
S.~Wang, J.~Zheng, H.-M. Hu, and B.~Li, ``Naturalness preserved enhancement
  algorithm for non-uniform illumination images,'' \emph{IEEE Transactions on
  Image Processing}, vol.~22, no.~9, pp. 3538--3548, 2013.

\bibitem{robust_retinex}
M.~Li, J.~Liu, W.~Yang, X.~Sun, and Z.~Guo, ``Structure-revealing low-light
  image enhancement via robust retinex model,'' \emph{IEEE Transactions on
  Image Processing}, vol.~27, no.~6, pp. 2828--2841, 2018.

\bibitem{WVM}
X.~Fu, D.~Zeng, Y.~Huang, X.-P. Zhang, and X.~Ding, ``A weighted variational
  model for simultaneous reflectance and illumination estimation,'' in
  \emph{Proceedings of the IEEE Conference on Computer Vision and Pattern
  Recognition}, 2016, pp. 2782--2790.

\bibitem{zhang2018high}
Q.~Zhang, G.~Yuan, C.~Xiao, L.~Zhu, and W.-S. Zheng, ``High-quality exposure
  correction of underexposed photos,'' in \emph{Proceedings of the 26th ACM
  international conference on Multimedia}, 2018, pp. 582--590.

\bibitem{zhang2019dual}
Q.~Zhang, Y.~Nie, and W.-S. Zheng, ``Dual illumination estimation for robust
  exposure correction,'' in \emph{Computer graphics forum}, vol.~38,
  no.~7.\hskip 1em plus 0.5em minus 0.4em\relax Wiley Online Library, 2019, pp.
  243--252.

\bibitem{ZeroShotRetinex}
A.~Zhu, L.~Zhang, Y.~Shen, Y.~Ma, S.~Zhao, and Y.~Zhou, ``Zero-shot restoration
  of underexposed images via robust retinex decomposition,'' in
  \emph{Proceedings of the IEEE International Conference on Multimedia and
  Expo}, 2020, pp. 1--6.

\bibitem{self_reinforced}
L.~Ma, R.~Liu, Y.~Wang, X.~Fan, and Z.~Luo, ``Low-light image enhancement via
  self-reinforced retinex projection model,'' \emph{IEEE Transactions on
  Multimedia}, pp. 1--1, 2022.

\bibitem{Untrained}
J.~Liang, Y.~Xu, Y.~Quan, B.~Shi, and H.~Ji, ``Self-supervised low-light image
  enhancement using discrepant untrained network priors,'' \emph{IEEE
  Transactions on Circuits and Systems for Video Technology}, vol.~32, no.~11,
  pp. 7332--7345, 2022.

\bibitem{pairlie}
Z.~Fu, Y.~Yang, X.~Tu, Y.~Huang, X.~Ding, and K.-K. Ma, ``Learning a simple
  low-light image enhancer from paired low-light instances,'' in
  \emph{Proceedings of the IEEE Conference on Computer Vision and Pattern
  Recognition}, 2023, pp. 22\,252--22\,261.

\bibitem{li2024rethinking}
D.~Li and S.~Rahardja, ``Rethinking affine transform for efficient image
  enhancement: A color space perspective,'' \emph{IEEE Transactions on
  Multimedia}, pp. 1--12, 2024.

\bibitem{liu2024pixel}
J.~Liu, Q.~Li, X.~Min, Y.~Su, G.~Zhai, and X.~Yang, ``Pixel-learnable 3dlut
  with saturation-aware compensation for image enhancement,'' \emph{IEEE
  Transactions on Multimedia}, vol.~26, pp. 11\,219--11\,231, 2024.

\bibitem{retinexnet}
C.~Wei, W.~Wang, W.~Yang, and J.~Liu, ``Deep retinex decomposition for
  low-light enhancement,'' \emph{arXiv preprint arXiv:1808.04560}, 2018.

\bibitem{retinexnet2}
W.~Yang, W.~Wang, H.~Huang, S.~Wang, and J.~Liu, ``Sparse gradient regularized
  deep retinex network for robust low-light image enhancement,'' \emph{IEEE
  Transactions on Image Processing}, vol.~30, pp. 2072--2086, 2021.

\bibitem{KinD}
Y.~Zhang, J.~Zhang, and X.~Guo, ``Kindling the darkness: A practical low-light
  image enhancer,'' in \emph{Proceedings of the 27th ACM International
  Conference on Multimedia}, 2019, pp. 1632--1640.

\bibitem{kind_plus}
Y.~Zhang, X.~Guo, J.~Ma, W.~Liu, and J.~Zhang, ``Beyond brightening low-light
  images,'' \emph{International Journal of Computer Vision}, vol. 129, pp.
  1013--1037, 2021.

\bibitem{RURS}
R.~Liu, L.~Ma, J.~Zhang, X.~Fan, and Z.~Luo, ``Retinex-inspired unrolling with
  cooperative prior architecture search for low-light image enhancement,'' in
  \emph{Proceedings of the IEEE Conference on Computer Vision and Pattern
  Recognition}, 2021, pp. 10\,556--10\,565.

\bibitem{uretinex}
W.~Wu, J.~Weng, P.~Zhang, X.~Wang, W.~Yang, and J.~Jiang, ``Uretinex-net:
  Retinex-based deep unfolding network for low-light image enhancement,'' in
  \emph{Proceedings of the IEEE Conference on Computer Vision and Pattern
  Recognition}, 2022, pp. 5901--5910.

\bibitem{HDRNet}
M.~Gharbi, J.~Chen, J.~T. Barron, S.~W. Hasinoff, and F.~Durand, ``Deep
  bilateral learning for real-time image enhancement,'' \emph{ACM Transactions
  on Graphics (TOG)}, vol.~36, no.~4, pp. 1--12, 2017.

\bibitem{DeepUPE}
R.~Wang, Q.~Zhang, C.-W. Fu, X.~Shen, W.-S. Zheng, and J.~Jia, ``Underexposed
  photo enhancement using deep illumination estimation,'' in \emph{Proceedings
  of the IEEE Conference on Computer Vision and Pattern Recognition}, 2019, pp.
  6842--6850.

\bibitem{sci}
L.~Ma, T.~Ma, R.~Liu, X.~Fan, and Z.~Luo, ``Toward fast, flexible, and robust
  low-light image enhancement,'' in \emph{Proceedings of the IEEE Conference on
  Computer Vision and Pattern Recognition}, 2022, pp. 5637--5646.

\bibitem{ZeroDCE}
C.~Guo, C.~Li, J.~Guo, C.~C. Loy, J.~Hou, S.~Kwong, and R.~Cong,
  ``Zero-reference deep curve estimation for low-light image enhancement,'' in
  \emph{Proceedings of the IEEE Conference on Computer Vision and Pattern
  Recognition}, 2020, pp. 1777--1786.

\bibitem{zerodce_plus}
C.~Li, C.~Guo, and C.~C. Loy, ``Learning to enhance low-light image via
  zero-reference deep curve estimation,'' \emph{IEEE Transactions on Pattern
  Analysis and Machine Intelligence}, vol.~44, no.~8, pp. 4225--4238, 2022.

\bibitem{luminancePyramid}
J.~Li, J.~Li, F.~Fang, F.~Li, and G.~Zhang, ``Luminance-aware pyramid network
  for low-light image enhancement,'' \emph{IEEE Transactions on Multimedia},
  vol.~23, pp. 3153--3165, 2021.

\bibitem{Laplacian}
S.~Lim and W.~Kim, ``Dslr: Deep stacked laplacian restorer for low-light image
  enhancement,'' \emph{IEEE Transactions on Multimedia}, vol.~23, pp.
  4272--4284, 2021.

\bibitem{wavelet}
J.~Xu, M.~Yuan, D.-M. Yan, and T.~Wu, ``Illumination guided attentive wavelet
  network for low-light image enhancement,'' \emph{IEEE Transactions on
  Multimedia}, pp. 1--14, 2022.

\bibitem{half_wavelet}
C.-M. Fan, T.-J. Liu, and K.-H. Liu, ``Half wavelet attention on m-net+ for
  low-light image enhancement,'' in \emph{Proceedings of the IEEE International
  Conference on Image Processing}, 2022, pp. 3878--3882.

\bibitem{attention-frequency}
Z.~He, W.~Ran, S.~Liu, K.~Li, J.~Lu, C.~Xie, Y.~Liu, and H.~Lu, ``Low-light
  image enhancement with multi-scale attention and frequency-domain
  optimization,'' \emph{IEEE Transactions on Circuits and Systems for Video
  Technology}, pp. 1--1, 2023.

\bibitem{EnlightenGAN}
Y.~Jiang, X.~Gong, D.~Liu, Y.~Cheng, C.~Fang, X.~Shen, J.~Yang, P.~Zhou, and
  Z.~Wang, ``Enlightengan: Deep light enhancement without paired supervision,''
  \emph{IEEE Transactions on Image Processing}, vol.~30, pp. 2340--2349, 2021.

\bibitem{Adversarial1}
Z.~Ni, W.~Yang, S.~Wang, L.~Ma, and S.~Kwong, ``Towards unsupervised deep image
  enhancement with generative adversarial network,'' \emph{IEEE Transactions on
  Image Processing}, vol.~29, pp. 9140--9151, 2020.

\bibitem{DPE}
Y.-S. Chen, Y.-C. Wang, M.-H. Kao, and Y.-Y. Chuang, ``Deep photo enhancer:
  Unpaired learning for image enhancement from photographs with gans,'' in
  \emph{2018 IEEE Conference on Computer Vision and Pattern Recognition}, 2018,
  pp. 6306--6314.

\bibitem{corss-disentanglement}
L.~Guo, R.~Wan, W.~Yang, A.~Kot, and B.~Wen, ``Cross-image disentanglement for
  low-light enhancement in real world,'' \emph{IEEE Transactions on Circuits
  and Systems for Video Technology}, pp. 1--1, 2023.

\bibitem{DRBN2}
W.~Yang, S.~Wang, Y.~Fang, Y.~Wang, and J.~Liu, ``Band representation-based
  semi-supervised low-light image enhancement: Bridging the gap between signal
  fidelity and perceptual quality,'' \emph{IEEE Transactions on Image
  Processing}, vol.~30, pp. 3461--3473, 2021.

\bibitem{flow}
Y.~Wang, R.~Wan, W.~Yang, H.~Li, L.-P. Chau, and A.~Kot, ``Low-light image
  enhancement with normalizing flow,'' in \emph{Proceedings of the AAAI
  Conference on Artificial Intelligence}, vol.~36, no.~3, 2022, pp. 2604--2612.

\bibitem{snr_transformer}
X.~Xu, R.~Wang, C.-W. Fu, and J.~Jia, ``Snr-aware low-light image
  enhancement,'' in \emph{Proceedings of the IEEE Conference on Computer Vision
  and Pattern Recognition}, 2022, pp. 17\,693--17\,703.

\bibitem{ultra_transformer}
T.~Wang, K.~Zhang, T.~Shen, W.~Luo, B.~Stenger, and T.~Lu,
  ``Ultra-high-definition low-light image enhancement: A benchmark and
  transformer-based method,'' in \emph{Proceedings of the AAAI Conference on
  Artificial Intelligence}, vol.~37, no.~3, 2023, pp. 2654--2662.

\bibitem{xu2024bilateral}
R.~Xu, Y.~Li, Y.~Niu, H.~Xu, Y.~Chen, and T.~Zhao, ``Bilateral interaction for
  local-global collaborative perception in low-light image enhancement,''
  \emph{IEEE Transactions on Multimedia}, vol.~26, pp. 10\,792--10\,804, 2024.

\bibitem{li2024sam}
G.~Li, B.~Zhao, and X.~Li, ``Low-light image enhancement with sam-based
  structure priors and guidance,'' \emph{IEEE Transactions on Multimedia},
  vol.~26, pp. 10\,854--10\,866, 2024.

\bibitem{wu2023_semantic}
Y.~Wu, C.~Pan, G.~Wang, Y.~Yang, J.~Wei, C.~Li, and H.~T. Shen, ``Learning
  semantic-aware knowledge guidance for low-light image enhancement,'' in
  \emph{Proceedings of the IEEE Conference on Computer Vision and Pattern
  Recognition}, 2023, pp. 1662--1671.

\bibitem{DRBN}
W.~Yang, S.~Wang, Y.~Fang, Y.~Wang, and J.~Liu, ``From fidelity to perceptual
  quality: A semi-supervised approach for low-light image enhancement,'' in
  \emph{Proceedings of the IEEE Conference on Computer Vision and Pattern
  Recognition}, 2020, pp. 3060--3069.

\bibitem{enc}
J.~Huang, Y.~Liu, X.~Fu, M.~Zhou, Y.~Wang, F.~Zhao, and Z.~Xiong, ``Exposure
  normalization and compensation for multiple-exposure correction,'' in
  \emph{Proceedings of the IEEE Conference on Computer Vision and Pattern
  Recognition}, 2022, pp. 6043--6052.

\bibitem{huang2022exposure}
J.~Huang, M.~Zhou, Y.~Liu, M.~Yao, F.~Zhao, and Z.~Xiong,
  ``Exposure-consistency representation learning for exposure correction,'' in
  \emph{Proceedings of the 30th ACM International Conference on Multimedia},
  2022, pp. 6309--6317.

\bibitem{fecnet}
J.~Huang, Y.~Liu, F.~Zhao, K.~Yan, J.~Zhang, Y.~Huang, M.~Zhou, and Z.~Xiong,
  ``Deep fourier-based exposure correction network with spatial-frequency
  interaction,'' in \emph{Proceedings of the European Conference on Computer
  Vision}.\hskip 1em plus 0.5em minus 0.4em\relax Springer, 2022, pp. 163--180.

\bibitem{wang2023decoupling}
Y.~Wang, L.~Peng, L.~Li, Y.~Cao, and Z.-J. Zha, ``Decoupling-and-aggregating
  for image exposure correction,'' in \emph{Proceedings of the IEEE Conference
  on Computer Vision and Pattern Recognition}, 2023, pp. 18\,115--18\,124.

\bibitem{cudi}
C.~Li, C.~Guo, R.~Feng, S.~Zhou, and C.~C. Loy, ``Cudi: Curve distillation for
  efficient and controllable exposure adjustment,'' \emph{arXiv preprint
  arXiv:2207.14273}, 2022.

\bibitem{huang2023learning}
J.~Huang, F.~Zhao, M.~Zhou, J.~Xiao, N.~Zheng, K.~Zheng, and Z.~Xiong,
  ``Learning sample relationship for exposure correction,'' in
  \emph{Proceedings of the IEEE Conference on Computer Vision and Pattern
  Recognition}, 2023, pp. 9904--9913.

\bibitem{cliplit}
Z.~Liang, C.~Li, S.~Zhou, R.~Feng, and C.~C. Loy, ``Iterative prompt learning
  for unsupervised backlit image enhancement,'' in \emph{Proceedings of the
  IEEE International Conference on Computer Vision}, 2023, pp. 8094--8103.

\bibitem{CoTF}
Z.~Li, F.~Zhang, M.~Cao, J.~Zhang, Y.~Shao, Y.~Wang, and N.~Sang, ``Real-time
  exposure correction via collaborative transformations and adaptive
  sampling,'' in \emph{Proceedings of the IEEE/CVF Conference on Computer
  Vision and Pattern Recognition}, 2024, pp. 2984--2994.

\bibitem{OSMamba}
G.~Li, B.~Chen, C.~Zhao, L.~Zhang, and J.~Zhang, ``Osmamba: Omnidirectional
  spectral mamba with dual-domain prior generator for exposure correction,'' in
  \emph{IEEE/CVF Conference on Computer Vision and Pattern Recognition
  (CVPR)}.\hskip 1em plus 0.5em minus 0.4em\relax IEEE, 2025, pp. 7480--7490.

\bibitem{CLIPRestoreX}
X.~Huang, Q.~Zhang, J.-F. Hu, and W.-S. Zheng, ``Clip-restorex: Restore image
  structure and perception in exposure correction,'' in \emph{Proceedings of
  the AAAI Conference on Artificial Intelligence}, vol.~39, no.~4.\hskip 1em
  plus 0.5em minus 0.4em\relax AAAI, 2025, pp. 3760--3768.

\bibitem{DDPM}
J.~Ho, A.~Jain, and P.~Abbeel, ``Denoising diffusion probabilistic models,''
  \emph{Advances in neural information processing systems}, vol.~33, pp.
  6840--6851, 2020.

\bibitem{DDIM}
J.~Song, C.~Meng, and S.~Ermon, ``Denoising diffusion implicit models,''
  \emph{arXiv preprint arXiv:2010.02502}, 2020.

\bibitem{DiffusionBeatGAN}
P.~Dhariwal and A.~Nichol, ``Diffusion models beat gans on image synthesis,''
  \emph{Advances in neural information processing systems}, vol.~34, pp.
  8780--8794, 2021.

\bibitem{CFG}
J.~Ho and T.~Salimans, ``Classifier-free diffusion guidance,'' \emph{arXiv
  preprint arXiv:2207.12598}, 2022.

\bibitem{LDM}
R.~Rombach, A.~Blattmann, D.~Lorenz, P.~Esser, and B.~Ommer, ``High-resolution
  image synthesis with latent diffusion models,'' in \emph{Proceedings of the
  IEEE/CVF conference on computer vision and pattern recognition}, 2022, pp.
  10\,684--10\,695.

\bibitem{lin2024diffbir}
X.~Lin, J.~He, Z.~Chen, Z.~Lyu, B.~Dai, F.~Yu, Y.~Qiao, W.~Ouyang, and C.~Dong,
  ``Diffbir: Toward blind image restoration with generative diffusion prior,''
  in \emph{European conference on computer vision}.\hskip 1em plus 0.5em minus
  0.4em\relax Springer, 2024, pp. 430--448.

\bibitem{ControlNet}
L.~Zhang, A.~Rao, and M.~Agrawala, ``Adding conditional control to
  text-to-image diffusion models,'' in \emph{Proceedings of the IEEE/CVF
  International Conference on Computer Vision}, 2023, pp. 3836--3847.

\bibitem{ExposureDiffusion}
Y.~Wang, Y.~Yu, W.~Yang, L.~Guo, L.-P. Chau, A.~C. Kot, and B.~Wen,
  ``Exposurediffusion: Learning to expose for low-light image enhancement,'' in
  \emph{IEEE/CVF International Conference on Computer Vision (ICCV)}.\hskip 1em
  plus 0.5em minus 0.4em\relax IEEE, 2023, pp. 12\,438--12\,448.

\bibitem{yi2023diff}
X.~Yi, H.~Xu, H.~Zhang, L.~Tang, and J.~Ma, ``Diff-retinex: Rethinking
  low-light image enhancement with a generative diffusion model,'' in
  \emph{Proceedings of the IEEE/CVF international conference on computer
  vision}, 2023, pp. 12\,302--12\,311.

\bibitem{sun2024dimensionx}
W.~Sun, S.~Chen, F.~Liu, Z.~Chen, Y.~Duan, J.~Zhang, and Y.~Wang, ``Dimensionx:
  Create any 3d and 4d scenes from a single image with controllable video
  diffusion,'' \emph{arXiv preprint arXiv:2411.04928}, 2024.

\bibitem{wang2025videoscene}
H.~Wang, F.~Liu, J.~Chi, and Y.~Duan, ``Videoscene: Distilling video diffusion
  model to generate 3d scenes in one step,'' in \emph{2025 IEEE/CVF Conference
  on Computer Vision and Pattern Recognition (CVPR)}.\hskip 1em plus 0.5em
  minus 0.4em\relax IEEE, 2025, pp. 16\,475--16\,485.

\bibitem{liu2024reconx}
F.~Liu, W.~Sun, H.~Wang, Y.~Wang, H.~Sun, J.~Ye, J.~Zhang, and Y.~Duan,
  ``Reconx: Reconstruct any scene from sparse views with video diffusion
  model,'' \emph{arXiv preprint arXiv:2408.16767}, 2024.

\bibitem{liu2024physics3d}
F.~Liu, H.~Wang, S.~Yao, S.~Zhang, J.~Zhou, and Y.~Duan, ``Physics3d: Learning
  physical properties of 3d gaussians via video diffusion,'' \emph{arXiv
  preprint arXiv:2406.04338}, 2024.

\bibitem{lv2024fourier}
X.~Lv, S.~Zhang, C.~Wang, Y.~Zheng, B.~Zhong, C.~Li, and L.~Nie, ``Fourier
  priors-guided diffusion for zero-shot joint low-light enhancement and
  deblurring,'' in \emph{Proceedings of the IEEE/CVF Conference on Computer
  Vision and Pattern Recognition}, 2024, pp. 25\,378--25\,388.

\bibitem{cho2024zero}
J.~Cho, S.~Aghajanzadeh, Z.~Zhu, and D.~Forsyth, ``Zero-shot low light image
  enhancement with diffusion prior,'' \emph{arXiv preprint arXiv:2412.13401},
  2024.

\bibitem{li2024sagiri}
B.~Li, S.~Ma, Y.~Zeng, X.~Xu, Y.~Fang, Z.~Zhang, J.~Wang, and K.~Chen,
  ``Sagiri: Low dynamic range image enhancement with generative diffusion
  prior,'' \emph{arXiv preprint arXiv:2406.09389}, 2024.

\bibitem{LIME}
X.~Guo, Y.~Li, and H.~Ling, ``Lime: Low-light image enhancement via
  illumination map estimation,'' \emph{IEEE Transactions on Image Processing},
  vol.~26, no.~2, pp. 982--993, 2017.

\bibitem{SID}
C.~Chen, Q.~Chen, J.~Xu, and V.~Koltun, ``Learning to see in the dark,'' in
  \emph{Proceedings of the IEEE Conference on Computer Vision and Pattern
  Recognition}, 2018, pp. 3291--3300.

\bibitem{adam}
D.~P. Kingma and J.~Ba, ``Adam: A method for stochastic optimization,''
  \emph{arXiv preprint arXiv:1412.6980}, 2014.

\bibitem{cosine}
I.~Loshchilov and F.~Hutter, ``Sgdr: Stochastic gradient descent with warm
  restarts,'' in \emph{International Conference on Learning Representations},
  2016.

\bibitem{ssim}
Z.~Wang, A.~C. Bovik, H.~R. Sheikh, and E.~P. Simoncelli, ``Image quality
  assessment: from error visibility to structural similarity,'' \emph{IEEE
  Transactions on Image Processing}, vol.~13, no.~4, pp. 600--612, 2004.

\bibitem{lpips}
R.~Zhang, P.~Isola, A.~A. Efros, E.~Shechtman, and O.~Wang, ``The unreasonable
  effectiveness of deep features as a perceptual metric,'' in \emph{Proceedings
  of the IEEE conference on computer vision and pattern recognition}, 2018, pp.
  586--595.

\bibitem{niqe}
A.~Mittal, R.~Soundararajan, and A.~C. Bovik, ``Making a “completely blind”
  image quality analyzer,'' \emph{IEEE Signal processing letters}, vol.~20,
  no.~3, pp. 209--212, 2012.

\end{thebibliography}








\end{document}